\documentclass[a4paper,fleqn]{cas-dc}
\usepackage{amsmath,amsfonts}
\usepackage{colortbl}
\usepackage{pifont}
\usepackage{textcomp}
\usepackage{stfloats}
\usepackage{threeparttable}
\usepackage{url}
\usepackage{verbatim}
\usepackage{graphicx}
\usepackage{diagbox}
\usepackage[ruled,vlined]{algorithm2e}
\usepackage{bm}
\usepackage{bbm}
\usepackage{soul}
\usepackage{subfig}
\usepackage{subfloat}
\usepackage{amsfonts}
\usepackage{multirow}
\usepackage{algpseudocode}
\usepackage{makecell}
\usepackage{booktabs}
\usepackage{xcolor}
\usepackage{booktabs}
\usepackage{bbding}
\usepackage{array}
\usepackage{color}
\definecolor{s1}{RGB}{117,159,204}
\definecolor{s2}{RGB}{189,208,233}
\usepackage[numbers]{natbib}

\def\tsc#1{\csdef{#1}{\textsc{\lowercase{#1}}\xspace}}
\tsc{WGM}
\tsc{QE}

\begin{document}
	\let\WriteBookmarks\relax
	\def\floatpagepagefraction{1}
	\def\textpagefraction{.001}
	
	\shorttitle{Towards Assessing the Synthetic-to-Measured Adversarial Vulnerability of SAR ATR}    
	
	\shortauthors{B.W. Peng, B. Peng, J. Xia, T. Liu, Y. Liu, L. Liu}  
	
	\title [mode = title]{Towards Assessing the Synthetic-to-Measured Adversarial Vulnerability of SAR ATR}  
	
	\author[1]{Bowen Peng}[orcid=0000-0002-6793-5025]
	
	\ead{pbow16@nudt.edu.cn}
	\affiliation[1]{organization={College of Electronic Science and Technology, National University of Defense Technology},
		city={Changsha},
		postcode={410073}, 
		state={Hunan},
		country={China}}	
	
	\author[2]{Bo Peng}
	\ead{ppbbo@nudt.edu.cn}
	
	\affiliation[2]{organization={Test Center, National University of Defense Technology},
		city={Xi'an},
		postcode={710100}, 
		state={Shaanxi},
		country={China}}
	
	\author[1]{Jingyuan Xia}
	\ead{j.xia10@nudt.edu.cn}
	
	\author[1]{Tianpeng Liu}
	\ead{everliutianpeng@sina.cn}
	
	\author[1]{Yongxiang Liu}
	\cormark[1]
	\ead{lyx_bible@sina.com}
	
	\author[1]{Li Liu}[orcid=0000-0002-2011-2873]
	\cormark[1]
	\ead{dreamliu2010@gmail.com}
	
	\cortext[1]{Corresponding author}

	\begin{abstract}	
		Recently, there has been increasing concern about the vulnerability of deep neural network (DNN)-based synthetic aperture radar (SAR) automatic target recognition (ATR) to adversarial attacks, where a DNN could be easily deceived by clean input with imperceptible but aggressive perturbations. 
		This paper studies the synthetic-to-measured (S2M) transfer setting, where an attacker generates adversarial perturbation based solely on synthetic data and transfers it against victim models trained with measured data. Compared with the current measured-to-measured (M2M) transfer setting, our approach does not need direct access to the victim model or the measured SAR data. We also propose the transferability estimation attack (TEA) to uncover the adversarial risks in this more challenging and practical scenario. The TEA makes full use of the limited similarity between the synthetic and measured data pairs for blind estimation and optimization of S2M transferability, leading to feasible surrogate model enhancement without mastering the victim model and data.
		Comprehensive evaluations based on the publicly available synthetic and measured paired labeled experiment (SAMPLE) dataset demonstrate that the TEA outperforms state-of-the-art methods and can significantly enhance various attack algorithms in computer vision and remote sensing applications. Codes and data are available at \url{https://github.com/scenarri/S2M-TEA}.
	\end{abstract}

	\begin{keywords}
		Synthetic aperture radar  \sep Automatic target recognition \sep Deep neural networks  \sep Adversarial attack \sep Transferability
	\end{keywords}
	
	\maketitle
	
	\section{Introduction}
	\label{}
	\textcolor{black}{As a longstanding, fundamental, and challenging problem
		in synthetic aperture radar (SAR) image interpretation, automatic target recognition (ATR)
		has been an active area of research for several decades \textcolor{black}{\cite{kechagias2021automatic,el2016automatic}}. The goal of SAR ATR is to determine the class labels of objects of interest (\emph{i.e.}, targets)~\textcolor{black}{\cite{dudgeon1993overview}}, and SAR ATR supports a variety of civilian and military applications, including modern airport management \textcolor{black}{\cite{airport}}, military and maritime surveillance (\emph{e.g.}, smuggling, piracy, or illegal fishing) \textcolor{black}{\cite{pawar2023sar,zhao2014ship}}, disaster alert \cite{8517531,liu2019contrario}, and rescue \cite{rashkovetsky2021wildfire}. In recent years, deep neural networks (DNNs), with their ability to automatically learn feature representations from data, have enabled significant progress in SAR ATR and emerged as the mainstream approach \cite{wang2023crucial,shi2024unsupervised,huang2024physics,zhang2020convolutional,zhang2020fec,li2021multiscale,zhang2021domain}.}
	
	\textcolor{black}{However, DNNs have inherent security vulnerabilities to adversarial attacks that can be exploited by adding deliberately crafted, human imperceptible perturbations to natural data that cause misclassifications \textcolor{black}{\cite{szegedy2013intriguing,harnessing2015goodfellow,kurakin2016adversarial}}. \textcolor{black}{Distinguished by utilizing different aspects of the victim models' information, adversarial attacks can be categorized into white-box, query-based, or transfer-based attacks.} \textcolor{black}{All victim model information, such as the architecture, weights, and gradient, is accessible in the white-box attack setting, and the adversarial perturbation can be generated by performing gradient ascent to maximize the classification loss function. In contrast, the query-based and transfer-based adversaries utilize the victim model's output or a surrogate model to complete the adversarial optimization process.} 
		These attacks present potential hazards for deployed DNN-based intelligent systems, and the hazards can be extreme in domains where security is critical, such as SAR ATR for military and maritime surveillance. Therefore, it is imperative to design \textcolor{black}{\cite{sva,serban2020adversarial,rsaa2}}, defend \textcolor{black}{\cite{9492037,ortiz2021optimism}, and understand \cite{xuRSsecurity,NEURIPS2019_e2c420d9,zhang2024does}} adversarial attack examples, and these examples serve as a surrogate to assess robustness and play a key role in developing more resilient DNN models for SAR ATR. } 
	Additionally, SAR ATR is an ideal area for studying adversarial risks, in part because there are many critical special requirements for deploying malicious examples against it. 
	For example, the high-stakes nature does not allow for cloud access or any white-box surrogate model to approximate the victim model's gradient. The unique imaging mechanism also requires special attention when designing perturbations to be physically injected into the imaging chain. Effective design of these perturbations requires detailed knowledge of the imaging geometries, the radiometric properties of targets and their surroundings, and the various radar operating parameters such as the imaging algorithms. 
	
\begin{figure*}[tbp]
		\centering
		\subfloat[Measured-to-measured attack with direct access to the victim model \textcolor{s1}{$\blacksquare$} (white-box methods  \cite{zhou2023attributed,peng2022scattering,fastcw,sarAAexperience2020}) or the use of measured training data to train the surrogate model \textcolor{s2}{$\blacksquare$} (M2M transfer-based methods  \cite{sva,lin2023boosting,sarAAempirical2021,xia2022sar,qin2023scma}) to calculate adversarial perturbations.]{\includegraphics[width=0.8\linewidth]{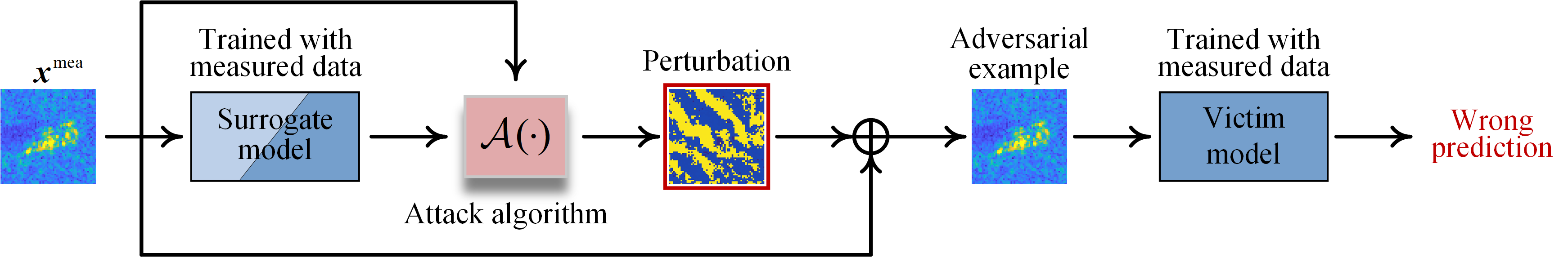}}\\
		
		\subfloat[Synthetic-to-measured attack where the surrogate model is trained on synthetic data, the victim model is trained on measured data, \textcolor{black}{and only the synthetic data is used in the attack algorithm.}]{\includegraphics[width=0.8\linewidth]{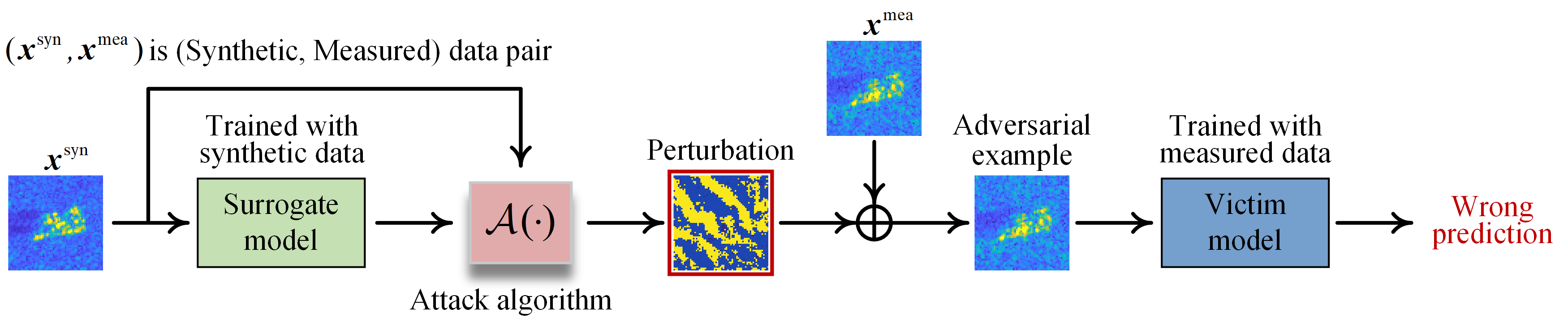}}
		\caption{Comparison between S2M and M2M attack settings.}
		\label{attackmodel}
	\end{figure*}

	\textcolor{black}{Currently, research on adversarial attacks in SAR ATR focuses on ensuring the practicality \cite{zhou2023attributed,qin2023scma,peng2022scattering,xia2022sar} or transferability \cite{sva,lin2023boosting} of adversarial examples.} Unfortunately, these studies typically focus on the victim model \cite{zhou2023attributed,peng2022scattering,fastcw,sarAAexperience2020} or its training data \cite{sva,lin2023boosting,sarAAempirical2021,xia2022sar,qin2023scma} to calculate effective adversarial examples\textcolor{black}{. In other words, current methods either directly access the victim model to perform white-box attacks or utilize measured victim model data to train a surrogate model for transfer-based attacks. We refer to these settings collectively as the measured-to-measured (M2M) setting, and this setting renders} the research insignificant or misleading since the real and practical adversarial risks are not available when the model and measured data are tightly protected. Therefore, we study the synthetic-to-measured (S2M) transfer setting in this work as a more realistic threat scenario. As shown in Fig. \ref{attackmodel}, an S2M adversary utilizes the knowledge about its own targets to synthesize SAR data \cite{franceschetti1992saras,lewis2019sar,kusk2016synthetic} for surrogate model training and victimizes a target model using perturbations crafted based on this synthetic data-trained surrogate model.

	Although a significant attack performance gap currently exists between S2M and the current state-of-the-art \textcolor{black}{measured-to-measured} (M2M) transfer setting\footnote{As an example, the best average attack success rate against eleven target models decreased from approximately 80\% to 40\% for M2M versus S2M, respectively, with a perturbation budget of $\epsilon=16/255$.}, we show this gap can be \textcolor{black}{narrowed without accessing} the measured data and victim model, revealing potential risks in the more practical S2M scenario. In particular, our purpose is to highlight the adversarial risks by improving the attack performance in the S2M setting. To that end, \textcolor{black}{we design an S2M transferability estimator and a model enhancement process to assimilate the gradient directions between the synthetic data-trained surrogate and the measured data-trained victim models without access to any of the measured data, and we refer to this as the transferability estimation attack (TEA). The S2M transferability estimator disentangles the gradient similarity between the surrogate and victim models to model and data discrepancies and serves as a substitute objective for blindly optimizing the surrogate's transferability.} We also demonstrate that a copy of the synthetic data with Gaussian noise can serve as a simple and effective solution to overcome these discrepancies and measure the S2M transferability with high quality. Furthermore, we modify the surrogate's architecture to expand a search space to acquire a higher transferability estimation while implicitly achieving better attack performance, and we provide new insights into the relationship between generalization and transferability from synthetic to measured data.
	
	\textcolor{black}{In summary, we provide insight into novel, transfer-based, black-box adversarial risks for DNN-based SAR ATR, and we show that even without direct access to the measured data, the S2M method can achieve non-negligible transfer attack performance against typical classifiers. Our work highlights the importance of dedicating resources to practical threat scenarios and securing ATR systems.} Overall, the main contributions of this paper are as follows:
	\begin{itemize} 
		\item 
		\textcolor{black}{To the best of our knowledge, this is the first work to study the S2M
			adversarial vulnerability of SAR ATR, \emph{i.e.},  the attack transferability of a surrogate model trained solely on synthetic data to a victim model trained on measured data.}
		
		\item We propose the TEA method and reveal the potential adversarial risks in the S2M setting. The TEA enables estimation of the S2M transfer attack capabilities and surrogate model enhancement without accessing the victim model and data. We also provide an effective blind parameter selection strategy to perform TEA.
		
		\item Through extensive evaluations involving a wide range of victim models and attack algorithms, we demonstrate that our estimator can effectively indicate the S2M transferability. We show that the TEA can significantly improve the S2M attack performance compared to various other approaches. We also show that our methods are compatible with various transferability-enhancing methods and the physical attacks in SAR ATR. 
		
		\item We show that the S2M with the TEA can effectively assess robustness in DNN-based SAR ATR systems, and we \textcolor{black}{freely offer the model weights and code of this work to promote further development.}
	\end{itemize}
	
	\textcolor{black}{The remainder of this paper is structured in the following manner. The background and related work is presented in Section \ref{relatedworks}.
		The details, application scenarios, and evaluation results of the S2M transfer setting are given in Section \ref{outmethod}. Section \ref{teasec} provides details of the TEA, including the S2M transferability estimator and the surrogate enhancement method, along with our parameter selection strategy. Our experimental setting and results are given in Section \ref{experiment}. We discuss the fundamental understanding and physical applicability of our method are in Section \ref{discussion}. The conclusion and plans for future work are provided in Section \ref{conclusion}.}
	
	\section{Background and related work}\label{relatedworks}
	Since attention was brought to neural network vulnerabilities more than a decade ago, there has been research dedicated to attacking and defending neural networks, and research on designing and defending adversarial examples has greatly contributed to the robustness and reliability of DNNs. This section provides a background on deep learning-based SAR ATR along with discussions on transfer-based attacks in computer vision and adversarial attacks in SAR ATR. 
	
	\subsection{\textcolor{black}{Deep learning-based SAR ATR}}
	Over the last decade, deep learning-based techniques have significantly impacted SAR ATR in target recognition performance with its automatic feature encoding and classification capabilities. \textcolor{black}{Since SAR ATR can be categorized as a subfield of computer vision, many off-the-shelf DNN models that are designed for optical image processing, such as ResNets \cite{resnet2016kh} and VGGNets \cite{vgg2015sk}, can be directly utilized and outperform conventional target recognition solutions, like sparse representation and scattering center-based methods
		\cite{shao2017performance,kechagias2021automatic}}. Despite initial success, researchers continue to pursue improvements in deep learning-based design methods for special requirements in SAR ATR, and one of the main focus areas is overcoming the difficulties associated with SAR data acquisition, such as lightweight design \cite{aconvnet2016chen,wang2022recognition}, insufficient data learning \cite{wang2023crucial,zhao2023few}, or target-background correlation elimination \cite{weijie,peng2023learning}. Another focus area is model design with SAR domain knowledge, such as the imaginary part of the data \cite{yu2019complex,zhang2021domain} and electromagnetic scattering information \cite{huang2024physics,li2021multiscale}.
	In this paper, we consider both advanced and lightweight models to investigate the performance of different DNN- and vision transformer-based methods.
	
	Synthetic data can also be utilized in SAR ATR, and the leading benchmark is the synthetic and measured paired labeled experiment (SAMPLE) dataset. This dataset provides matched synthetic-measured data pairs and has assisted development in various techniques, such as transfer learning \cite{shi2024unsupervised,malmgren2017improving}, synthetic-measured transformation \cite{lewis2018generative}, and data augmentation \cite{sellers2020augmenting}. These techniques can help bridge the gap between synthetic and measured SAR data, allowing for more effective and practical recognition tasks, and the work most closely related to ours is generalizing a model trained with solely synthetic data to correctly recognize the measured data \cite{inkawhich2021bridging}. 
	
	\subsection{Transfer-based attacks in computer vision}\label{transferattackrelatedwork}
	\subsubsection{Problem formulation}
	\textcolor{black}{Target recognition in computer vision involves input images, $\bm{x}\in\mathcal{X}$, along with their corresponding labels, $y\in\mathcal{Y}$, and a well-trained classifier, $f:\mathcal{X}\rightarrow\mathcal{Y}$, is responsible for predicting labels for the given inputs. An adversary aims to falsify the classifier prediction with an imperceptible yet powerful perturbation, $\bm{\delta}$, that satisfies
		\begin{equation}
			f(\bm{x}+\bm{\delta})\neq y \quad \text{s.t.} \quad \mathcal{D}(\bm{x}+\bm{\delta}, \bm{x})\leq\epsilon.
		\end{equation}
		Here, the function $\mathcal{D(\cdot)}$ measures a distance and cooperates with the perturbation budget, $\epsilon$, to ensure stealthiness, or imperceptibility. The attack objective is usually transferred as maximization of the cross-entropy, $\mathcal{L}_{\text{CE}}$, while restricting $\bm{\delta}$ within an $\epsilon$-bounded $l_{\infty}$-ball as
		\begin{equation}\label{attackgoal}
			\underset{\bm{\delta}}{\operatorname{maximize}} \,\, \mathcal{L}_{\text{CE}}(f(\bm{x}+\bm{\delta}), y) \quad \text{s.t.} \quad \|\bm{\delta}\|_{\infty}\leq\epsilon,
		\end{equation}
		where $\bm{\delta}$ can be generated by various attack algorithms, $\mathcal{A}(\cdot)$, depending on a given attack setting.}
	
	There has been a considerable amount of work dedicated to enhancing the transferability of transfer-based attacks, and in this section, we categorize the mainstream attack methods into algorithmic methods and surrogate-side methods.
	
	\subsubsection{Algorithmic methods}
	In this paper, we limit our scope of algorithmic methods to gradient-based, generative, and universal attacks. With gradient-based transfer attacks, one typically utilizes a surrogate model trained on the same dataset as the target victim model, and the perturbation is generated via gradient ascent. With the widely adopted distance constraint that restricts $\bm{\delta}$ within an $\epsilon$-bounded $l_{\infty}$-ball, the plain gradient-based attack can be summarized as
	\begin{equation}\label{}
		\bm{x}_{0}^{\text{adv}} = \bm{x}, \quad \bm{x}_{i+1}^{\text{adv}} = \bm{x}_{i}^{\text{adv}} + \alpha\cdot\operatorname{Sign}(\nabla_{\bm{x}}\mathcal{L}_{\text{CE}}(f(\bm{x}_{i}^{\text{adv}}), y)),
	\end{equation}
	where $\alpha$ represents the step size, \textcolor{black}{$\operatorname{Sign}(\cdot)$ is the Signum function} and $\bm{\delta}=\bm{x}^{\text{adv}}-\bm{x}$. Considerable efforts have been made to enhance the transferability of gradient-based attacks, \textcolor{black}{and these efforts can be divided into advanced optimization methods and input transformation-based methods.} The first category includes many advanced gradient calculation methods, such as the momentum iterative (MI) attack method \cite{momentum2018dong}, the Nesterov iterative (NI) attack method \cite{nestrov2019lin}, and the variance tuning (VT) attack method \cite{Wang_2021_CVPR}, to overcome the issue of getting trapped in local optima. The second category includes methods that calculate gradients on the image(s) transformed by label-preserving transformations, such as the diversity input (DI) attack method \cite{inputdiversity2019xie}, the scale-invariant (SI) attack method \cite{nestrov2019lin}, and the translation-invariant (TI) attack method \cite{evading2019dong}. 
	
	Generative attacks train a generator by attacking the surrogate model over a set of data points \cite{poursaeed2018generative}, and after training, the generator should be able to effectively deceive the system as it receives unfamiliar data points and target models. The relative cross-entropy loss \cite{CDA2019} and intermediate features are commonly utilized to pursue better transferability \cite{zhang2021beyond,LTAP2021}. It is also possible to optimize a single universal perturbation that can effectively attack a diverse range of SAR images and target models \cite{peng2022empirical}. With model parameters frozen, the universality can be achieved by optimizing $\bm{\delta}$ to maximize classification loss \cite{domainfeatureuap}, diversify the original output \cite{datafreeuap}, or ignite spurious features \cite{GDUAP} in mini-batch training over a large amount of data points.
	
	\subsubsection{Surrogate-based methods}
	In addition to pursuing better attack capability in optimization algorithms, research has also been dedicated to refining the surrogate model \cite{wu2019skip,guo2020backpropagating,zhang2021backpropagating,ZhuDRA,yang2022boosting}. In one example, the distribution-relevant attack (DRA) method fine-tunes the surrogate model to align the gradient direction with the conditional data distribution density \cite{ZhuDRA}. The dark surrogate model (DSM) method utilizes the soft output of a surrogate model to train a more transferable one \cite{yang2022boosting}, while the little robust surrogate (LRS) method uses adversarial examples with a little perturbation budget to train a surrogate model \cite{springer2021little}.
	One important branch of surrogate refinement methods is structural modification, where previous studies have provided substantial evidence highlighting the significant impact of activation functions and skip connections on model transferability. \textcolor{black}{For example, linear backpropagation (LinBP) \cite{guo2020backpropagating} and continuous backpropagation (ConBP) \cite{zhang2021backpropagating} backpropagate the gradients more linearly or smoothly compared with the rectified linear unit (ReLU) function}, which enhances the transferability. Furthermore, the skip gradient method (SGM) \cite{wu2019skip} and the intrinsic adversarial attack (IAA) method \cite{zhu2021rethinking} have revealed that the ratio of the residual module to the skip connection plays a crucial role in both the accuracy and transferability of the model. These findings emphasize the importance of considering the design and configuration of model architecture when trying to enhance surrogate transferability. 
	
	\subsection{Adversarial attack in SAR ATR}\label{attackinsar}
	Following the pioneering works in 2020 \cite{zhiweireview,xuRSsecurity}, there has been a surge of research interest in exploring the adversarial vulnerability of DNN-based SAR ATR models, and early on, researchers focused on proving and evaluating the vulnerability and characteristics, leading to many valuable observations. For example, researchers found that SAR ATR models exhibit similar vulnerability to optical models in white-box attack settings, and the wrong predictions of adversarial SAR images seem to follow a specific distribution related to the object structure \cite{sarAAempirical2021,peng2022scattering}. Researchers have also directed attention toward understanding the domain characteristics of radar countermeasures, including applicability and transferability. 
	
	\textcolor{black}{To date, several attempts have been made to design adversarial examples with physical constraints, such as manipulating the location \cite{zhou2023attributed} or other attributes \cite{qin2023scma} of existing scattering centers or appending additional adversarial scatterers \cite{peng2022scattering}. The implementation of digital perturbations in an electromagnetic environment has also been explored using existing jamming tools \cite{xia2022sar}. From the transferability side, researchers have suggested that manipulating the speckle noise \cite{sva} or intermediate features \cite{lin2023boosting} could provide better transfer attack performance and highlight the adversarial risks.} However, the main body of current research on transferability follows the M2M transfer setting, and the experiments generally train surrogate and target models using the same data distribution. In this work, we investigate the inadequacy of this setting and assess the adversarial vulnerability of SAR ATR with the S2M setting. It is worth noting that our work is compatible with studies that focus on physical applicability by providing them with reference digital adversarial examples of better transferability (see Section \ref{physicalapplicability}).

	\section{\textcolor{black}{The proposed S2M transfer attack setting}}\label{outmethod}
	In this section, we present some unique aspects of the SAR ATR that must be accounted for when considering adversarial attacks, such as the creation, feasibility, and application scenarios of adversarial examples regarding the attacker side and the victim side.
	
	\subsection{The S2M transfer setting}\label{ourattackmodel}
	\subsubsection{Current attack settings in SAR ATR}
	\textcolor{black}{Existing attack approaches for SAR ATR either utilize the victim model itself or train a surrogate with the training data of the victim model to generate adversarial perturbations, such as training the surrogate and victim models using the same training set of the MSTAR dataset \cite{mstar}.} These methods also directly access the target's measured data when evaluating the victim model's robustness to give
	\begin{equation}\label{currentsetting}
		\bm{\delta}=\mathcal{A}(\textcolor{black}{f^{\text{tar}/\text{mea}}}, \textcolor{black}{\bm{x}^{\text{mea}}}, y, \epsilon),
	\end{equation}
	where $f^{\text{tar}/\text{mea}}$ represents the exact target victim model or a surrogate model trained with the same data distribution and $\bm{x}^{\text{mea}}$ is the measured data point to be attacked. 
	
	\subsubsection{The S2M setting}
\begin{figure}[tbp]
		\centering
		\includegraphics[width=0.85\linewidth]{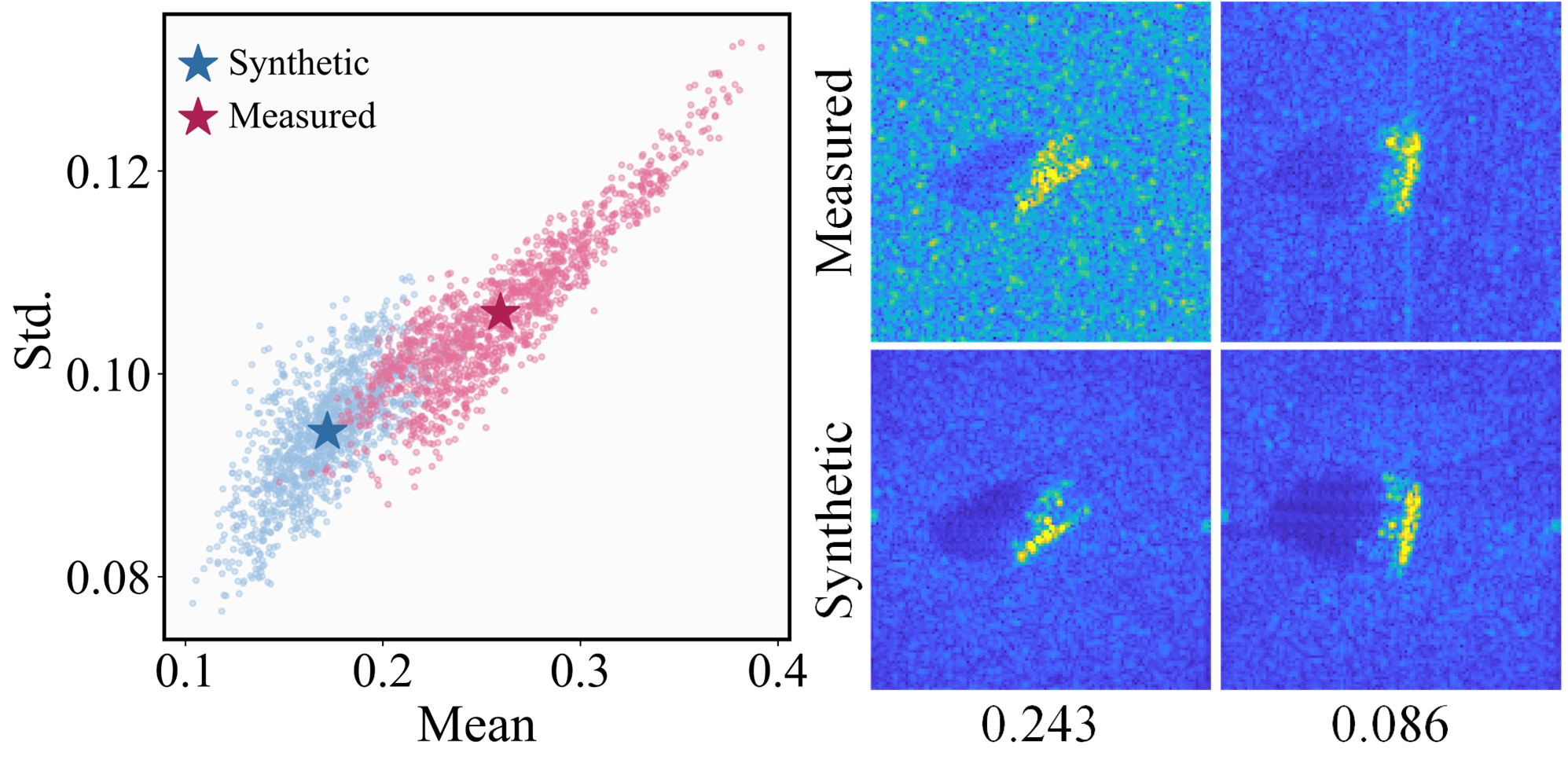}
		\caption{Differences between the synthetic and measured data of the SAMPLE dataset: (\textbf{Left}) the mean value and standard deviation (Std.) and (\textbf{Right}) the paired instances with the lowest and highest root mean squared error.}
		\label{datadiff}
	\end{figure}

	We contend that the above setting is inappropriate in the field of SAR ATR since the victim model and the measured data are generally inaccessible. To improve the current M2M approach for the practical scenario where measured data is unavailable, we propose the S2M setting. In this setting, we use a surrogate model trained with synthetic data to assess the vulnerability of a  target model trained with measured data, allowing perturbation to be generated with synthetic data and transferred to attack the measured data. Specifically, we consider an attacker attempting to victimize a target SAR ATR model, $f^\text{tar}$, that has been trained on a measured SAR dataset, $\bm{x}^\text{mea} \in \mathcal{X}^\text{mea}$, deployed by the victim. The attacker holds a paired synthetic dataset\footnote{Here, we assume the synthetic and measured data are paired one-by-one for simplicity. Section \ref{s2mvariantions} considers the unpaired scenario.}, $\bm{x}^\text{syn} \in \mathcal{X}^\text{syn}$, that allows it to train a surrogate model, $f^\text{sur}$, for transfer attack. The goal is consistent with Eq. \eqref{attackgoal}, and then the perturbation is generated based on a given attack algorithm, surrogate model, synthetic data, and perturbation budget as
	\begin{equation}\label{attacksetting2}
		\bm{\delta}=\mathcal{A}(f^{\text{sur}}, \bm{x}^{\text{syn}}, y, \epsilon).
	\end{equation}
	Recall the comparison between S2M and M2M illustrated in Fig. \ref{attackmodel}. An obvious difference between S2M and M2M is that an S2M adversary trains the surrogate model and generates adversarial perturbation only using the synthetic data, but it should be noted that the synthetic and measured data are not perfectly matched due to various factors, \textcolor{black}{such as limitations in electromagnetic calculations, the data processing schedule, and the imaging algorithm}. The gap between S2M and M2M is shown in Fig. \ref{datadiff} using the SAMPLE dataset as an example, where it is clear that S2M is a more challenging setting than M2M for an attacker. 
	
	\subsubsection{Application scenarios}
	A suitable attack setting alerts the victim to where the operating chain would be maliciously utilized and evaluates model robustness along with potential defense strategies. The proposed S2M suits defense purposes by building an appropriate adversary, where full knowledge of a targeted image (\emph{i.e.}, the electromagnetic structure of targets and the statistics prior of surroundings) is readily mastered by a potential attacker. The other main pieces of information required to synthesize SAR data \cite{franceschetti1992saras,lewis2019sar,kusk2016synthetic}, such as viewing geometry, radar frequency, and resolution, can be obtained by analyzing intelligence and received signals. This information could include the direction of arrival estimation in the case of airborne SAR and the orbital elements in the case of satellite SAR. For evaluation and defense, S2M can be used directly to test models and design potential defense methods, and S2M manifests as an attack type that has not been investigated by the SAR ATR community, encouraging new defense approaches and advancing the understanding between synthetic and measured data. \textcolor{black}{In this context, all information about our own targets, radar, and ATR algorithms is available to construct the synthetic dataset and surrogate model for strong potential adversaries}. 
	
	Another natural question is whether the adversarial perturbations could be injected into the SAR system to ignite real threats, and if not, there is no point in researching the transfer risks. Building on the information in Section \ref{attackinsar}, we provide a detailed discussion on this topic in Section \ref{physicalapplicability}.
	
	\subsection{Evaluation}\label{S2MEVAL}
	In this section, we report a preliminary comparison between the S2M and M2M transfer attack settings. We trained eleven target models with the measured data of the SAMPLE \cite{lewis2019sar} dataset including a ResNet-18 model. Another ResNet-18 model was trained with the synthetic data of the dataset. Table \ref{comparabbandS2M} reports the average attack success rate (ASR, see Eq. \eqref{asrdef}) against the target models achieved by these two ResNet-18 surrogates with six representative transfer-based attacks, and the table shows the degradation in ASR for S2M compared to M2M. More experimental details are provided in Section \ref{experiment}. The best result in the M2M  transfer scenario was 79.53\% achieved by cross-domain attack (CDA) compared to 42.50\% achieved by TI in the S2M transfer scenario, and the ASR for all attack algorithms degrade by more than 40\% with S2M compared to M2M. Clearly, a surrogate trained with the same data distribution as the targets yields significant benefits to the attacker, achieving satisfactory attack performance. However, attacking SAR ATR models in this manner is not feasible due to the lack of access to both the data and the model. More detailed experimental settings comparing S2M and M2M are presented in Section \ref{experiment}. 
	
	\begin{table}[htbp]
		\centering
		\caption{Average ASR (\%) against eleven target models trained over the measured dataset with a ResNet-18 surrogate model trained over the synthetic (S2M) and measured (M2M) datasets and a perturbation budget of $16/255$ for normalized data. The performance degradation ($\frac{ASR_{\text{M2M}}-{ASR_{\text{S2M}}}}{ASR_{\text{M2M}}}\times 100\%$) is included with the S2M results.}
		\begin{tabular}{rcc}
			\toprule
			& \multicolumn{2}{c}{Transfer scenario} \\ \cmidrule{2-3}
			\makecell[c]{Attack} & \makecell[c]{Measured$\rightarrow$Measured\\(M2M)} & \makecell[c]{Synthetic$\rightarrow$Measured\\(S2M)} \\ \hline
			PGD \cite{madry2018towards} & 51.87 & 24.97$_{51.86\%\downarrow}$\\
			TI \cite{evading2019dong} & 78.67 & 42.50$_{45.98\%\downarrow}$\\
			CDA \cite{CDA2019} & 79.53  & 39.50$_{50.33\%\downarrow}$\\
			BIA \cite{zhang2021beyond} & 79.01 & 41.09$_{47.99\%\downarrow}$\\
			DF-UA \cite{domainfeatureuap} & 47.48 & 28.15$_{40.71\%\downarrow}$\\
			CS-UA \cite{datafreeuap} & 47.99 & 25.69$_{46.47\%\downarrow}$\\
			\bottomrule
		\end{tabular}
		\label{comparabbandS2M}
	\end{table}
	
	\section{Transferability estimation-based S2M attacks \textcolor{black}{e}nhancement}\label{teasec}
	
	The observed performance gap between S2M and M2M encourages us to enhance the S2M transferability to better reveal and assess the adversarial risks of SAR ATR for surrogate models trained using synthetic data. In this section, we present the TEA, which consists of an estimator that can blindly mirror the S2M transferability and an estimation-guided surrogate enhancement process. The enhanced surrogate holds promise in powering various existing attack algorithms in the S2M setting, and an overview of the TEA algorithm is summarized in Algorithm \ref{algoritm} with details explained in the following subsections.
	
	\subsection{Motivation}
	We aim to highlight the adversarial vulnerability by performing aggressive attacks under the challenging S2M setting. As discussed in Section \ref{transferattackrelatedwork}, transfer-based attacks are broadly categorized into gradient-based, generative, and universal methods. In this paper, we focus on enhancing the gradient-based methods, but note that our approach also shows satisfactory effectiveness for other attack methods. In gradient-based methods, an adversary attacks the target model, $f^\text{tar}$, on dataset $\bm{x}^\text{mea}$ w.r.t. label $y$ by the gradient using the surrogate model based on the synthetic substitute dataset $\bm{x}^\text{syn}$, denoted as $\nabla_{\bm{x}}\mathcal{L}_{\text{CE}}(f^\text{sur}(\bm{x}^\text{syn}), y)$. Consequently, for $(\bm{x}^\text{mea}, \bm{x}^\text{syn}) \sim (\mathcal{X}^\text{mea}, \mathcal{X}^\text{syn})$, our objective is to make
	\begin{equation}\label{eq1}
		\nabla_{\bm{x}}\mathcal{L}_{\text{CE}}(f^\text{sur}(\bm{x}^\text{syn}), y)\approx\nabla_{\bm{x}}\mathcal{L}_{\text{CE}}(f^\text{tar}(\bm{x}^\text{mea}), y).
	\end{equation}
	Since the measured data and target model are inaccessible, we aim to enhance our surrogate model to achieve better gradient similarity, as an enhanced model can strengthen a variety of attack algorithms. Meanwhile, since most attacks generate perturbations based on the \textit{ascending direction} of the gradient, we can relax the objective as maximization of the cosine similarity (\textit{CosSim}) as:
	\begin{equation}\label{eq2}
		\underset{\bm{\Theta},\bm{\Lambda}}{\operatorname{maximize}} \,\, \mathbb{E}_{\mathcal{X}^\text{syn}, \mathcal{X}^\text{mea}}\left[\textit{CosSim}(\nabla_{\bm{x}}\mathcal{L}_{\bm{\Theta},\bm{\Lambda}}^\text{sur}(\bm{x}^\text{syn}), \nabla_{\bm{x}}\mathcal{L}^\text{tar}(\bm{x}^\text{mea}))\right].
	\end{equation}
	Here, $\bm{\Theta}$ represents the model weights, $\bm{\Lambda}$ represents the architecture hyper-parameters (\textit{e.g.}, hyper-parameters for the activation function and skip connections), and $\nabla_{\bm{x}}\mathcal{L}_{\bm{\Theta},\bm{\Lambda}}^\text{sur}(\bm{x}^\text{syn})$ is an abbreviation of $\nabla_{\bm{x}}\mathcal{L}_{\text{CE}}(f^\text{sur}_{\bm{\Theta},\bm{\Lambda}}(\bm{x}^\text{syn}), y)$ for simplicity. Unfortunately, optimizing objective \eqref{eq2} is still not feasible in the S2M setting. Therefore, we devise a substitute estimator to measure the transferability of S2M for optimization purposes.
	
	\subsection{S2M transferability estimator}\label{estimator}

	\subsubsection{Transferability estimator}
	Starting with Eq. \eqref{eq2}, we disentangle the discrepancy between the two gradients of different models w.r.t. different datasets into two parts: 1) data discrepancy and 2) model discrepancy. To address the data discrepancy, we introduce a substitute dataset, $\mathcal{X}^\text{sub}$, and aim to enhance the transferability and generalization of our surrogate model, $f^{\ast\text{sur}}$, on this dataset. This is achieved by maximizing the following loss on $\mathcal{X}^\text{sub}$
	\begin{equation}\label{eq3}
		\mathcal{L}_{\text{Data}} =  \textit{CosSim}(\nabla_{\bm{x}}\mathcal{L}^{\ast\text{sur}}(\bm{x}^\text{sub}), \nabla_{\bm{x}}\mathcal{L}^{\ast\text{sur}}(\bm{x}^\text{syn})).
	\end{equation} 
	By maximizing $\mathcal{L}_{\text{Data}}$, our surrogate can better align the gradient directions between paired data points $(\bm{x}^{\text{syn}}, \bm{x}^{\text{sub}})$. If the substitute dataset is of good quality, meaning $\mathcal{X}^\text{sub} \approx \mathcal{X}^\text{mea}$, $f^{\ast\text{sur}}$ can effectively enhance the transferability of the surrogate model, allowing it to leverage its gradient to attack the target model using only synthetic data, and
	formally, $\mathcal{L}_{\text{Data}}$ indicates the model's transferability against the substitute dataset. However as $\mathcal{L}_{\text{Data}}$ increases, the surrogate model may become corrupted in terms of its performance on the original task of $\mathcal{X}^\text{syn}\rightarrow\mathcal{X}^\text{mea}$ when $\mathcal{X}^\text{sub}$ fails to accurately simulate the measured data. Unfortunately, this is a common occurrence since there is generally very limited knowledge available for the measured data.
	
	\begin{figure}[tbp]
		\centering
		\includegraphics[width=0.9\linewidth]{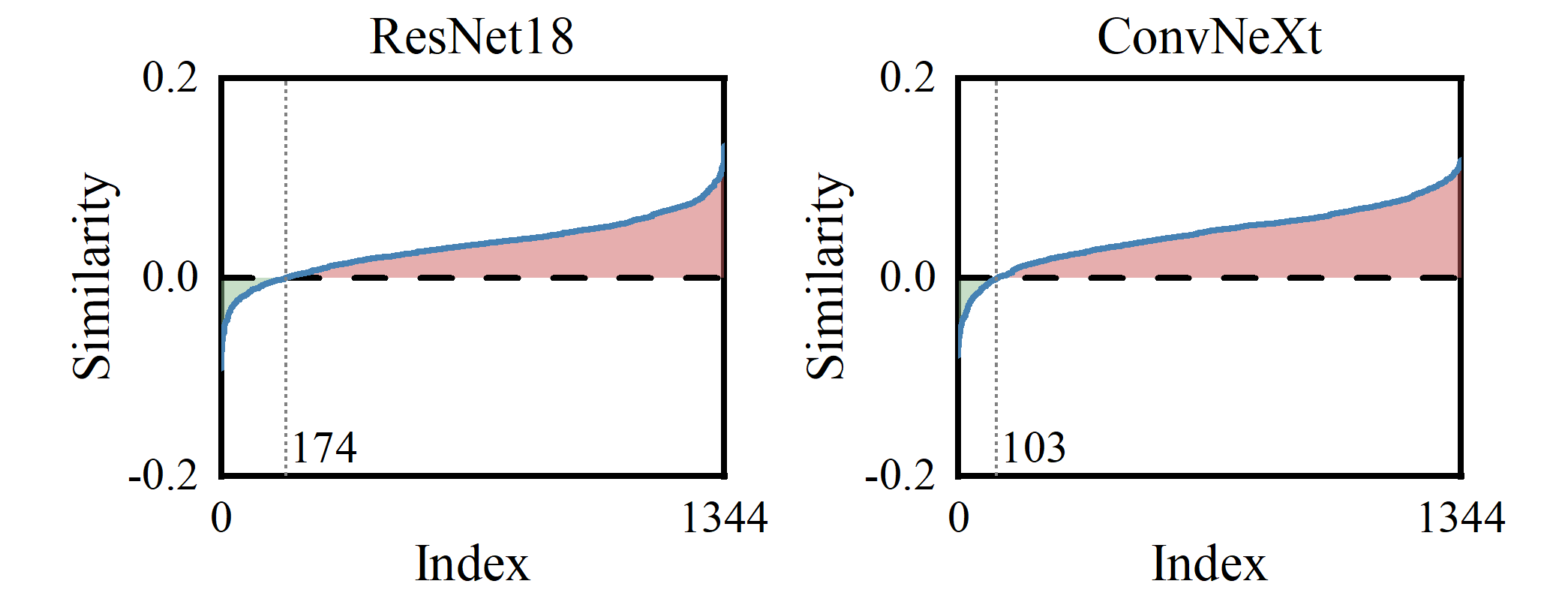}
		\caption{The average cosine similarity between gradient directions of the surrogate model (ResNet-18 and ConvNeXt) and eleven target models over 1345 synthetic-measured image pairs of the SAMPLE dataset in ascending order. The directions were calculated using projected gradient descent (PGD) attack \cite{madry2018towards}, and the first positive index is labeled in each plot.} 
		\label{datasimilarity}
	\end{figure}
	Regarding model discrepancy, the surrogate model is expected to generate gradient directions similar to the target model for the same input, which also cannot be explicitly measured. Here, we tackle both the model discrepancy and the limitation of $\mathcal{L}_{\text{Data}}$ together. 
	In particular, we assume an intrinsic similarity between the synthetic and measured data, a result of the data coming from the same electromagnetic structure, also results in a \textit{subtle intrinsic similarity} between the gradient direction of the surrogate and victim models. This similarity would not provide sufficiently effective transferability, but it is significant enough to be exploited. The empirical evidence is outlined in Fig. \ref{datasimilarity}, where the average cosine similarity between the gradient directions of the surrogate model and eleven target models are shown. We leverage this similarity to enhance the surrogate along the track of the original datasets and avoid neglecting the domain knowledge of $\mathcal{X}^{\text{syn}}$ and $\mathcal{X}^{\text{mea}}$. This is equivalent to building a conditional process that enhances the transferability while guaranteeing its robustness against $\mathcal{X}^{\text{mea}}$ from $\mathcal{X}^{\text{syn}}$. In practice, we pursue better alignment on the gradient direction between $f^{\ast\text{sur}}$ and $f^{\text{sur}}$ as:
	\begin{equation}\label{eq4}
		\mathcal{L}_{\text{Model}} =  \textit{CosSim}(\nabla_{\bm{x}}\mathcal{L}^{\ast\text{sur}}(\bm{x}^\text{syn}), \nabla_{\bm{x}}\mathcal{L}^\text{sur}(\bm{x}^\text{syn})).
	\end{equation} 
	Finally, we composite $\mathcal{L}_{\text{Data}}$ and $\mathcal{L}_{\text{Model}}$ to indicate the S2M transferability of a surrogate model against the target models with:
	\begin{equation}\label{eq5}
		\mathcal{L}_{\text{Total}} = \frac{1}{2}(\mathcal{L}_{\text{Data}} +  \mathcal{L}_{\text{Model}}).
	\end{equation}
	\textcolor{black}{We choose equal weighting since to the two measurements are in the same scale, and we do not further fine-tune the ratio so that we do not violate the inaccessibility of target models and measured data.}
	
	\subsubsection{Substitute data selection}
	At this point, the transferability would be ideally estimated through $\mathcal{L}_{\text{Total}}$ if the substitute data sufficiently matches the measured one. \textcolor{black}{However, due to the lack of sufficient knowledge about the measured data, we simply utilize the synthetic data with additive noise as the substitute}:
	\begin{equation}\label{eq6}
		\bm{x}^{\text{sub}} = \bm{x}^{\text{syn}} + \bm{n},  \text{where} \, \, \bm{n} \sim \mathcal{N}(0, \sigma^{2}),
	\end{equation}
	\textcolor{black}{where $\mathcal{N}(0, \sigma^{2})$ represents the zero-mean Gaussian distribution with a standard deviation of $\sigma$ which controls the distance from synthetic to substitute data. Note that the above estimation is reasonable when we posit the synthetic, measured, and substitute data is all derived from the same electromagnetic structure.}
	\begin{figure*}[htbp]
		\centering
		\includegraphics[width=1\linewidth]{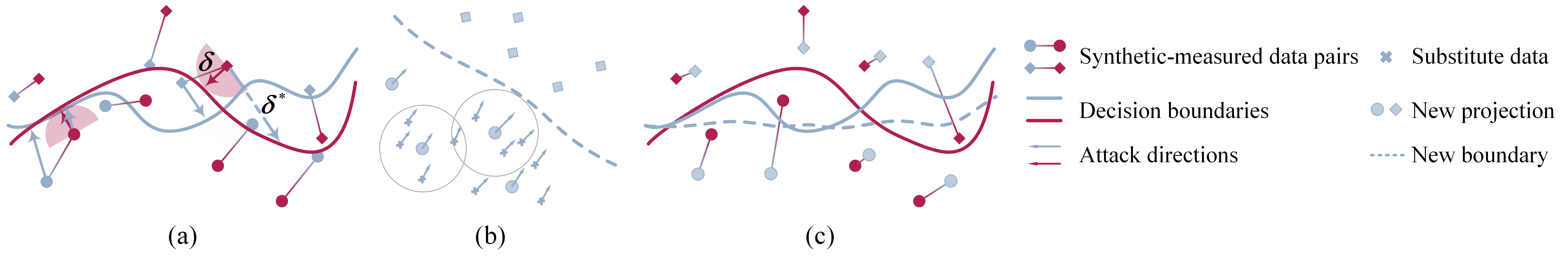}
		\caption{A simple schematic diagram of our estimator from a feature distribution perspective: (a) The data projections and decision boundaries of the surrogate and target model, where $\bm{\delta}$ and $\bm{\delta}^{\ast}$ indicate the minimum perturbation strength for a successful attack of white-box and S2M transfer attacks, respectively. (b) Optimizing $\mathcal{L}_{\text{data}}$ provides a flatter surrogate decision boundary, as it may not always be effective in fitting the original distribution and neglects the intrinsic similarity.  (c) Cooperation with  $\mathcal{L}_{\text{Model}}$ to optimize the total estimation leads to a smoother boundary and a new surrogate that retains the original distribution.}
		\label{framework}
	\end{figure*}
	This statistical substitute opens the door to a new perspective in understanding the proposed estimator $\mathcal{L}_{\text{Total}}$. As shown in Fig. \ref{framework}(a), the subtle similarity (\textit{i.e.}, positive similarities for most of the data pairs) leads to approximately similar feature projections, while a negative correlation requires a much larger perturbation budget to perform a successful attack. To this end, more similarity in these feature distributions and a flatter decision boundary help orient the synthetic gradient to the average direction of measured data. Furthermore, strengthening $\mathcal{L}_{\text{Data}}$ aligns gradient directions over the neighborhood of each data projection and leads to a smoother decision boundary while the new boundary may deviate from the original distributions, as illustrated by Fig. \ref{framework}(b). As a solution, we can pursue the smoothness while memorizing the original distribution by combining $\mathcal{L}_{\text{Data}}$ and $\mathcal{L}_{\text{Model}}$ as the optimization objective. We also notice the above analysis aligns well with the up-to-date theoretical understanding of adversarial transferability \cite{zhang2024does}, and further discussion is provided in Section \ref{understanding}.
	
	\subsection{Estimator-guided surrogate enhancement}\label{TEA}
	
	\begin{algorithm}[tb]
		\caption{Transferability estimation attack\label{algoritm}}
		\KwIn{Surrogate model, $f_{\bm{\Theta}, \bm{\Lambda}}^\text{sur}$; synthetic dataset,$\mathcal{X}^\text{syn}$, and labels, $\mathcal{Y}$; weight factor, $\lambda$; standard deviation, $\sigma$, for substitute data; attack algorithm, $\mathcal{A}(\cdot)$; perturbation budget, $\epsilon$; maximum epochs for FT, $N$; learning rate, $\eta$}
		\KwOut{Enhanced surrogate model, $f_{\bm{\Theta}^\ast, \bm{\Lambda}^\ast}^\text{sur+FT+AS}$; a set of adversarial perturbations, $\{\bm{\delta}\}$}
		$\bm{\rhd}$ \textbf{Fine-tuning} \\
		$\bm{\Theta}_0\gets\bm{\Theta}$ \\
		\For(\quad Eq. \eqref{FT}: fine-tuning weights by mini-batch training){$i\gets1$ \text{to} $N$}
		{\For(){$\mathcal{B} \sim (\mathcal{X}^{\text{syn}},\mathcal{Y})$} 
			{$\bm{\Theta}_i\gets\bm{\Theta}_{i-1}-\eta\nabla_{\bm{\Theta}}\mathcal{L}_{\text{FT}, \bm{\Theta}_{i-1}}(\mathcal{B})$}} 
		$\bm{\Theta}^\ast\gets\bm{\Theta}_N$\\
		$\bm{\rhd}$ \textbf{Architecture selection} \\
		Solve Eq. \eqref{AS} by Bayes optimization to find $\bm{\Lambda}^\ast\gets[\beta^\ast,\xi^\ast]$. \\
		Obtain $	f^{\text{sur+FT+AS}}_{\bm{\Theta}^\ast, \bm{\Lambda}^\ast}$ by replacing ReLU with $\operatorname{Softplus}_{\beta^\ast}$ and insert decay factor $\xi^\ast$ for skip connections.\\
		$\bm{\rhd}$ \textbf{Obtaining perturbations} \\
		$\bm{\delta}\gets\{\}$\\
		\For(){$(\bm{x}^{\text{syn}}_{i}, y_{i}) \in (\mathcal{X}^{\text{syn}},\mathcal{Y})$} 
		{$\{\bm{\delta}\}\gets\{\bm{\delta}, \mathcal{A}(f^{\text{sur+FT+AS}}_{\bm{\Theta}^\ast, \bm{\Lambda}^\ast}, \bm{x}^{\text{syn}_{i}}, y_{i}, \epsilon)\}$}
		\textbf{return} $f^{\text{sur+FT+AS}}_{\bm{\Theta}^\ast, \bm{\Lambda}^\ast}, \{\bm{\delta}\}$;
	\end{algorithm}
	
	Using Eq. \eqref{eq5} to estimate the transferability of the synthetic-measured model, we can proceed to identify a suitable surrogate model for S2M, and building upon the previous analysis, we develop a two-stage estimator-guided surrogate enhancement process.
	
	The two stages of the TEA method are fine-tuning (FT) and architecture selection (AS), and these stages are designed with consideration for the possibility of overfitting or other issues that could affect the accuracy of the $\mathcal{L}_{\text{Total}}$ as a good estimate of transferability. Therefore, we adopt a sequential approach where we first perform fine-tuning to improve generalization and obtain better initial weights, and we then enhance the model by searching for architecture hyper-parameters that yield higher values of $\mathcal{L}_{\text{Total}}$. This two-stage arrangement allows us to mitigate potential problems and optimize the overall performance of the model.
	
	In the initial stage, we begin with a pre-trained model, $f^{\text{sur}}$, trained on synthetic dataset, and to \textcolor{black}{obtain better initial weights for AS}, we fine-tune the model's weights using $\mathcal{L}_{\text{Data}}$. We can then write the FT loss function as:
	\begin{equation}
		\mathcal{L}_{\text{FT}} =  \mathcal{L}_{\text{CE}}(\bm{x}^{\text{syn}})-\lambda\mathcal{L}_{\text{Data}}(\bm{x}^{\text{syn}}, \bm{x}^{\text{sub}}),
	\end{equation}
	where $\lambda>0$ controls the weight of $\mathcal{L}_{\text{Data}}$. The fine-tuning process can be formulated as \textcolor{black}{minimization of the expectation of $\mathcal{L}_{\text{FT}}$ over synthetic dataset}:
	\begin{equation}\label{FT}
		\bm{\Theta}^\ast \gets \underset{\bm{\Theta}}{\operatorname{argmin}} \,\, \mathbb{E}_{\mathcal{X}^\text{syn},\mathcal{N}(0, \sigma) }\left[\mathcal{L}_{\text{FT},\bm{\Theta},\bm{\Lambda}}\right].
	\end{equation}
	Here, we specify the inputs of the above abbreviation as $\mathcal{L}_{\text{FT}}(f^{\text{sur}}_{\bm{\Theta},\bm{\Lambda}},\bm{x}^{\text{syn}}, \sigma,y)$ to avoid ambiguity.
	
	In the second stage, we aim to further exploit the potential of $f^{\text{sur+FT}}_{\bm{\Theta}^\ast, \bm{\Lambda}}$ by investigating different model architectures. Specifically, we search for architecture hyper-parameters, $\bm{\Lambda}$, that result in higher values of the $\mathcal{L}_{\text{Total}}$	metric. This can be formulated as:
	\begin{equation}\label{AS}
		\bm{\Lambda}^\ast \gets \underset{\bm{\Lambda}}{\operatorname{argmax}} \,\, \mathbb{E}_{\mathcal{X}^\text{syn},\mathcal{N}(0, \sigma) }\left[\mathcal{L}_{\text{Total},\bm{\Theta}^\ast,\bm{\Lambda}}\right].
	\end{equation}
	At this stage, unlike model training or fine-tuning, the model parameters, such as the weights and bias of the convolution kernel, are fixed. Inspired by recent studies, we define the search space for the activation function and skip connections, and we solve the above process using Bayesian optimization.
	
	\begin{figure}[tbp]
		\centering
		\includegraphics[width=1\linewidth]{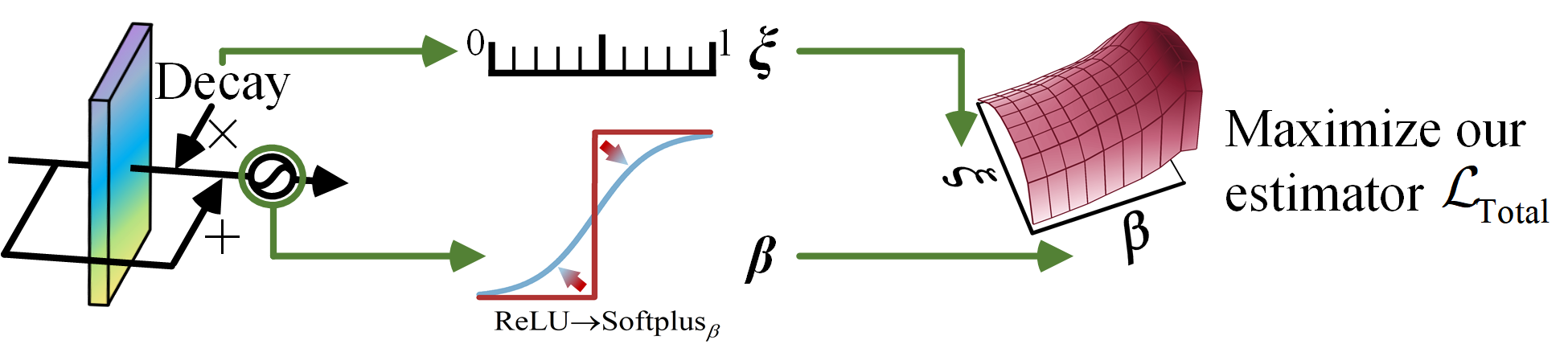}
		\caption{Process to construct the search space for AS showing a single layer as an example. Note that the figure shows the first derivatives for the ReLU function and the $\operatorname{Softplus}_{\beta}$ function.}
		\label{ASillustration}
	\end{figure}
	
	We modify the activation functions and the skip connections to construct the search space for AS according to the process outlined in Fig. \ref{ASillustration}. First, the $\operatorname{Softplus}_{\beta}$ activation is introduced as a smooth substitute for the widely used ReLU activation: \begin{equation}\label{eq8}\operatorname{Softplus}_\beta(x)=\frac{1}{\beta} \log (1+\exp (\beta x)).
	\end{equation}
	Second, we insert a decay factor, $\xi$, into the skip connections, which are widely deployed in the residual blocks of ResNet-like models:
	\begin{equation}\label{softplus}
		f_{i+1}(\bm{x}) = f_{i}(\bm{x}) + \xi_{i} \cdot g_{i}(f_{i}(\bm{x})), \quad 0 < \xi_{i} < 1,
	\end{equation}
	\textcolor{black}{where $g_{i}(\cdot)$ is the residual module at layer $i$}. For simplicity, we set a single decay factor for all skip connections, and our search space for AS is: 
	\begin{equation}\label{insertdecay}
		\bm{\Lambda}^\ast \in \{\bm{\Lambda}|\bm{\Lambda}=[\beta, \xi], 0 < \beta < 10, 0 < \xi < 1 \},
	\end{equation} 
	where the upper bound for $\beta$ is chosen from experiments where a level trend of $\mathcal{L}_{\text{Total}}$ is detected. \textcolor{black}{To be more specific, we changed $\beta$ and calculated the value of $\mathcal{L}_{\text{Total}}$, and the change in $\mathcal{L}_{\text{Total}}$ was no longer significant when $\beta>10$.} The TEA searches for the maximum $\mathcal{L}_{\text{Total}}$ over the two hyper-parameters, $\xi$ and $\beta$, to determine the S2M transferability estimation. Comparison between our method and related approaches \cite{zhang2021backpropagating,zhu2021rethinking,wu2019skip} is provided in Section \ref{comSOTA}.
	
	\subsection{\textcolor{black}{Parameter selection strategy}}\label{PSstrategy}
	In the previous subsections, we outlined the TEA that enhances a surrogate model for better performance in the S2M transfer setting, and here, we provide the parameter selection strategy for blind optimization in the absence of access to the target model and measured data. The FT is a model training procedure that involves selections for training epochs, learning rate, $\lambda$, and $\sigma$, resulting in a very large parameter space to investigate. To effectively fine-tune the surrogate, we set a long enough training timeline for fine-tuning that includes several instances of learning rate decay. 
	Intuitively, a larger value of $\sigma$ will result in lower values of $\mathcal{L}_{\text{Data}}$ and $\mathcal{L}_{\text{Total}}$ when the model weights are fixed, and a smaller value of $\sigma$ will result in larger values of $\mathcal{L}_{\text{Data}}$ and $\mathcal{L}_{\text{Total}}$. Furthermore, excessively large or small values of $\mathcal{L}_{\text{Total}}$ may not accurately indicate transferability due to the saturation of cosine similarity. Therefore, we choose the value of $\sigma_{\text{FT}}$ at which the surrogate model achieves $\mathcal{L}_{\text{Data}}$ of approximately 0.5 in the FT stage. For AS, we select $\sigma_{\text{AS}}$ at which the fine-tuned surrogate model achieves $\mathcal{L}_{\text{Total}}$ in the range of 
	0.2 to 0.5,  which allows for a relatively large positive variance in the $\mathcal{L}_{\text{Total}}$ value to facilitate the optimization process. We evaluate the effectiveness of our strategy in Section \ref{sensitivity}.

	\begin{figure*}[htbp]
		\centering
		\includegraphics[width=1\linewidth]{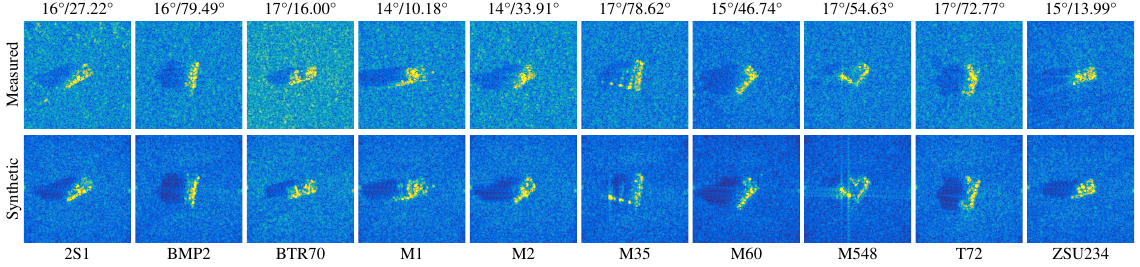}
		\caption{Examples of the synthetic and measured paired data in the SAMPLE dataset:  (\textbf{Top}) measured images where the heading indicates the azimuth/elevation angle and (\textbf{Bottom}) paired synthetic images.}
		\label{datasetfig}
	\end{figure*}
	\begin{table}[tbp]
		\centering
		\caption{Details of the SAMPLE data used in our experiments.}
		\begin{tabular}{cccc}
			\toprule
			Category & Seria \# & \# Synthetic & \# Measured	\\
			\midrule
			2S1 & B01 & 177 & 177\\
			BMP2 &9563 & 108 & 108 \\
			BTR70 & C71 & 96 & 96 \\
			M1 & 0AP00N & 131 & 131 \\
			M2 & MV02GX & 129 & 129 \\
			M35 & T839 & 131 & 131 \\
			M60 & C245HAB & 129 & 129 \\
			M548 & 3336 & 178 & 178 \\
			T72 & 812 & 110 & 110 \\
			ZSU234 & D08 & 177 & 177 \\
			\midrule
			Total & & 1345 & 1345 \\
			\bottomrule
		\end{tabular}
		\label{datasetconfiguration}
	\end{table}
	\section{Experiments}\label{experiment}
	In our experiments, the goal was to evaluate our method without utilizing any measured data for parameter selection. Therefore, we performed evaluations with our parameter selection strategy reported in Section \ref{PSstrategy} and analyzed the parameter sensitivity of the TEA. The following subsections provide details of the experimental setups.
	
	\subsection{Setup}\label{setting}
	\subsubsection{Dataset}
	The SAMPLE dataset \cite{lewis2019sar} was publicized\footnote{\url{https://github.com/benjaminlewis-afrl/SAMPLE_dataset_public}} by the Air Force Research Laboratory (AFRL) to facilitate synthetic data-assisted SAR ATR that could be generalized to various scenarios. The dataset consists of 1345 synthetic-measured data pairs of ten vehicle target categories with instances illustrated in Fig. \ref{datasetfig} and details outlined in Table \ref{datasetconfiguration}. The data is arranged in $128\times128$ pixels, covering azimuth angles from $10^{\circ}$ to $80^{\circ}$ and elevation angles from $14^{\circ}$ to $17^{\circ}$. 
	
	\begin{table}[tbp]
		\centering	
		\caption{Number of parameters, FLOPs (calculated for a single $64\times64$ input), accuracy for measured data (\%), and year the model was introduced for the studied DNN models.}
		\begin{tabular}{rcccc}
			\toprule
			\makecell[c]{Model} & \# Params. & FLOPs & Accuracy  & Year
			\\ \midrule
			ACN \cite{aconvnet2016chen} & 1.18$\times10^{5}$ & 8.91$\times10^{6}$ & 100.00 & 2016 \\
			SNV2 \cite{Ma_2018_ECCV} & 3.52$\times10^{5}$ & 3.12$\times10^{6}$ & 100.00 & 2018 \\
			MNV2 \cite{mobilev2} & 2.24$\times10^{6}$ & 2.61$\times10^{7}$ & 100.00 & 2018 \\
			RGN \cite{radosavovic2020designing} & 3.91$\times10^{6}$ & 3.41$\times10^{7}$ & 100.00 & 2020 \\
			EN \cite{efficientnet} & 4.02$\times10^{6}$ & 3.38$\times10^{7}$ & 99.78 & 2019 \\
			DN121 \cite{densenet2017huang} & 6.96$\times10^{6}$ & 2.30$\times10^{8}$ & 100.00 & 2017 \\
			RN18 \cite{resnet2016kh} & 1.12$\times10^{7}$ & 2.85$\times10^{8}$ & 100.00 & 2016 \\
			SwinT \cite{liu2021swin} & 1.89$\times10^{7}$ & 2.42$\times10^{8}$ & 100.00 & 2021 \\
			CNX \cite{liu2022convnet} & 2.78$\times10^{7}$ & 3.64$\times10^{8}$ & 100.00 & 2022 \\
			ViT \cite{dosovitskiy2020image} & 2.84$\times10^{7}$ & 1.85$\times10^{9}$ & 100.00 & 2020 \\
			VGG16 \cite{vgg2015sk} & 1.34$\times10^{8}$ & 2.74$\times10^{9}$ & 100.00 & 2015 \\
			\bottomrule
			\label{models}
		\end{tabular}
	\end{table}
	
	\subsubsection{Models}
	To better measure the transferability, we investigated a total of eleven target models, including AConvNet (ACN) \cite{aconvnet2016chen}, ShuffleNetV2 x0.5 (SNV2) \cite{Ma_2018_ECCV}, MobileNetV2 (MNV2) \cite{mobilev2}, RegNet y\_400mf (RGN) \cite{radosavovic2020designing}, EfficientNet-B0 (EN) \cite{efficientnet}, DenseNet-121 (DN121) \cite{densenet2017huang}, ResNet-18 (RN18) \cite{resnet2016kh}, Swin Transformer swin\_t (SwinT) \cite{liu2021swin}, ConvNeXt tiny (CNX) \cite{liu2022convnet}, Vision Transformer vit\_b\_16 (ViT) \cite{dosovitskiy2020image}, and VGG-16 \cite{vgg2015sk}. Specifics of these target models are listed in Table \ref{models}. We trained three surrogate models based on the synthetic dataset using RN18, RN34, and CNX. All surrogate models achieved accuracy levels of more than 99.9\% for synthetic data, and the accuracy levels for RN18, RN34, and CNX were 66.47\%, 58.29\%, and 45.58\%, respectively, for measured data.
	
	\subsubsection{Implementation details}
	We used all available data for training due to the limited amount of data, and the single-channel data was center-cropped to $64\times64$ and normalized to $[0,1]$ for training \cite{lewis2019sar,inkawhich2021bridging}. No other data augmentation techniques were utilized, and all eleven target models and three surrogate models were trained using the stochastic gradient descent (SGD) optimizer (with a momentum of 0.9 and weight decay of 0.0001) and cross-entropy loss. We searched for an appropriate initial learning rate within $\{0.01, 0.005, 0.001\}$ for each model and decayed it by 0.2 at the 20th and 30th epochs during a total of 50 training epochs.
	
	In the experiments, we performed FT on the synthetic data-trained RN18, RN34, CNX models using the SGD optimizer and $\mathcal{L}_{\text{FT}}$ loss, and we used $\sigma_{\text{FT}}$ values of 0.2, 0.2, and 0.25 for RN18, RN34, and CNX, respectively,  with $\lambda=1$ for 20 epochs. The initial learning rate was 0.005 and decayed by 0.2 at the 10th and 15th epochs. We solved the AS process using the gp\_minimize function from \textit{scikit-optimize}\footnote{\url{https://scikit-optimize.github.io/stable/}}, which involved 10 random starts and a total of 50 calls \cite{zhu2021rethinking}, and unless otherwise specified, all attacks were conducted under a perturbation budget of $\epsilon=16/255$. All gradient-based attacks were equipped with sign projection \cite{qin2021adversarial}, and the iteration was set to 10 with $\alpha=\epsilon/8$.
	
	\subsubsection{Comparison metric}
	The ASR was defined to measure the transferability of surrogate models. Specifically, given a surrogate and an attack algorithm, we generate adversarial perturbations $\{\bm{\delta}\}$ for all the synthetic data, and then a target model is tested with the attacked measured data. The ASR is then calculated as:
	\begin{equation}
		\textit{ASR}=[\sum_{i} \mathbb{I} ( f^{\text{tar}}(\bm{x}^{\text{mea}}_{i}\!+\!\bm{\delta}_{i}) \!\neq\! y_{i})/|\mathcal{X}^{\text{mea}}|] \times 100\%,
		\label{asrdef}
	\end{equation}
	where $\mathbb{I}(\cdot)$ represents the indicator function. In our experiments, we used the average ASR against the eleven target models to indicate the \textit{S2M transferability} of the given surrogate model and an attack algorithm, and with the same attack algorithm, a higher average ASR indicates better transferability of a surrogate model.
	
	\subsection{Effectiveness of the TEA}\label{evaluation}
	
	In this section, we utilize the RN18 to show the effectiveness of the TEA, including the S2M transferability estimator and the model-enhancing process.
	
	\begin{figure}[tbp]
		\centering	
		\includegraphics[width=0.8\linewidth]{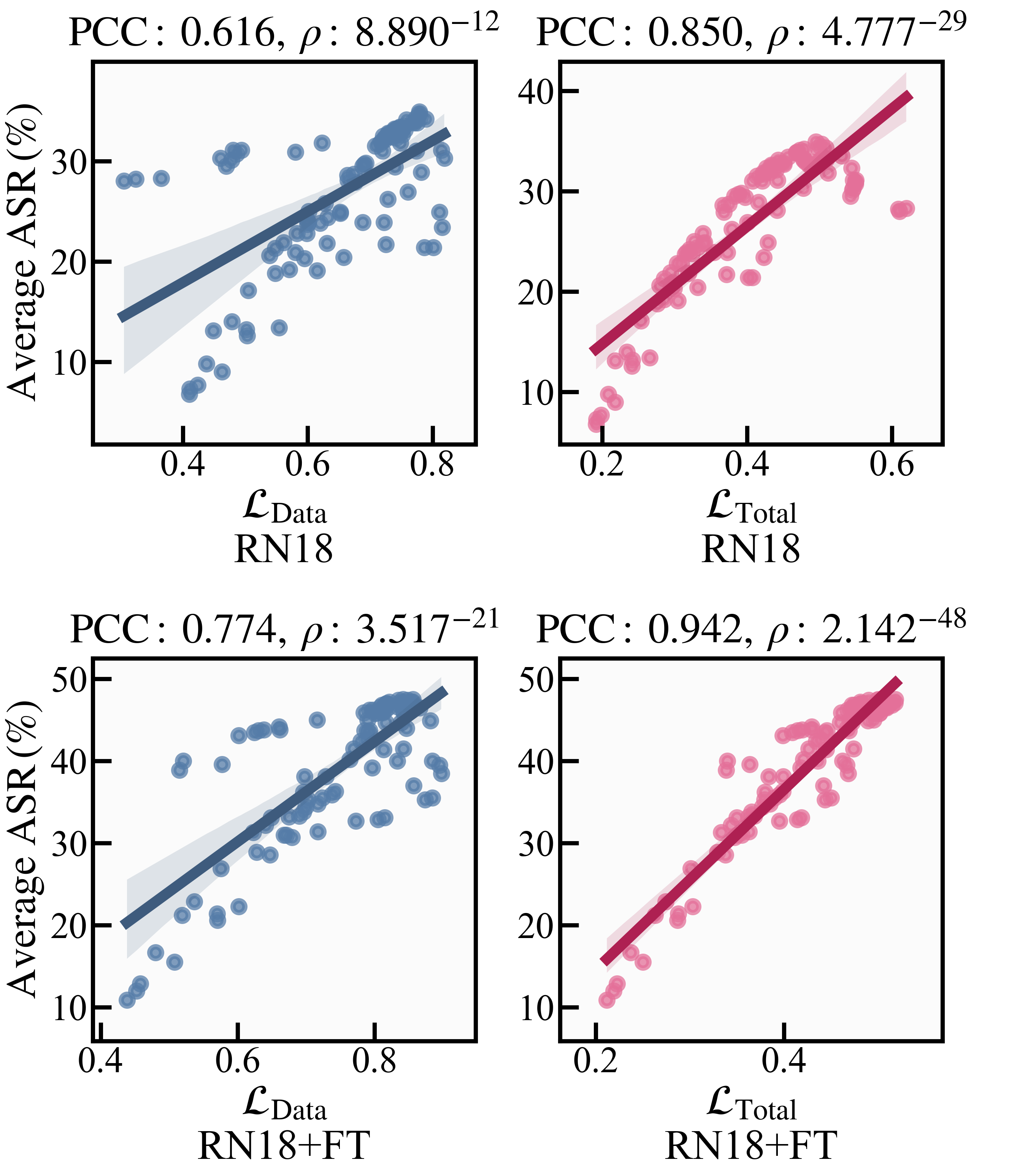}
		\caption{Results of Pearson correlation test, where the data was obtained with \textcolor{black}{the same random architecture hyper-parameters} to test the correlation of the average ASR with $\mathcal{L}_{\text{Data}}$ and $\mathcal{L}_{\text{Total}}$.}
		\label{pearson}
		\vspace{-4pt}
	\end{figure} 
	
	\subsubsection{Effectiveness of the transferability estimator}
	We first verified the quality of our S2M transferability estimator using the Pearson correlation test with one hundred combinations of the architecture hyper-parameters uniformly sampled from $\beta\sim U(0.5,10.5)$ and $\xi\sim U(0,1)$, for RN18 and its FT-enhanced version. Fig. \ref{pearson} shows the Pearson correlation coefficients (PCCs) and $\rho$-values for the average ASR versus $\mathcal{L}_{\text{Data}}$ and $\mathcal{L}_{\text{Total}}$. The results demonstrate that the proposed estimator can indicate the S2M transferability for both the original and FT-enhanced RN18, as $\mathcal{L}_{\text{Total}}$ achieved a PCC of 0.850 for RN18 and a PCC 0.942 for RN18+FT. Moreover, the higher PCC values for RN18+FT (0.850 \textit{vs.} 0.616 and 0.942 \textit{vs.} 0.774) show the effectiveness of the $\mathcal{L}_{\text{Model}}$ as an additional constraint on $\mathcal{L}_{\text{Data}}$.
	
	\subsubsection{Effectiveness of the FT enhancement}
	The data in Fig. \ref{pearson} also shows that with FT enhancement, the surrogate models exhibited stronger correlations with $\mathcal{L}_{\text{Total}}$ (0.942 \textit{vs.} 0.850 in PCC and 2.142$^{-48}$ \textit{vs.} 4.777$^{-29}$ in $\rho$-value). The data also shows that the model performance (mean of average ASRs) and potential (maximum of average ASRs) are simultaneously improved through FT enhancement. Therefore, FT renders the AS process more efficient and effective in finding well-performing architecture hyper-parameters, which validates the appropriateness of the sequential order of FT	and AS processes in TEA.
	
	\subsubsection{Effectiveness of the AS enhancement}
\begin{figure}[tbp]
		\centering
		\includegraphics[width=1\linewidth]{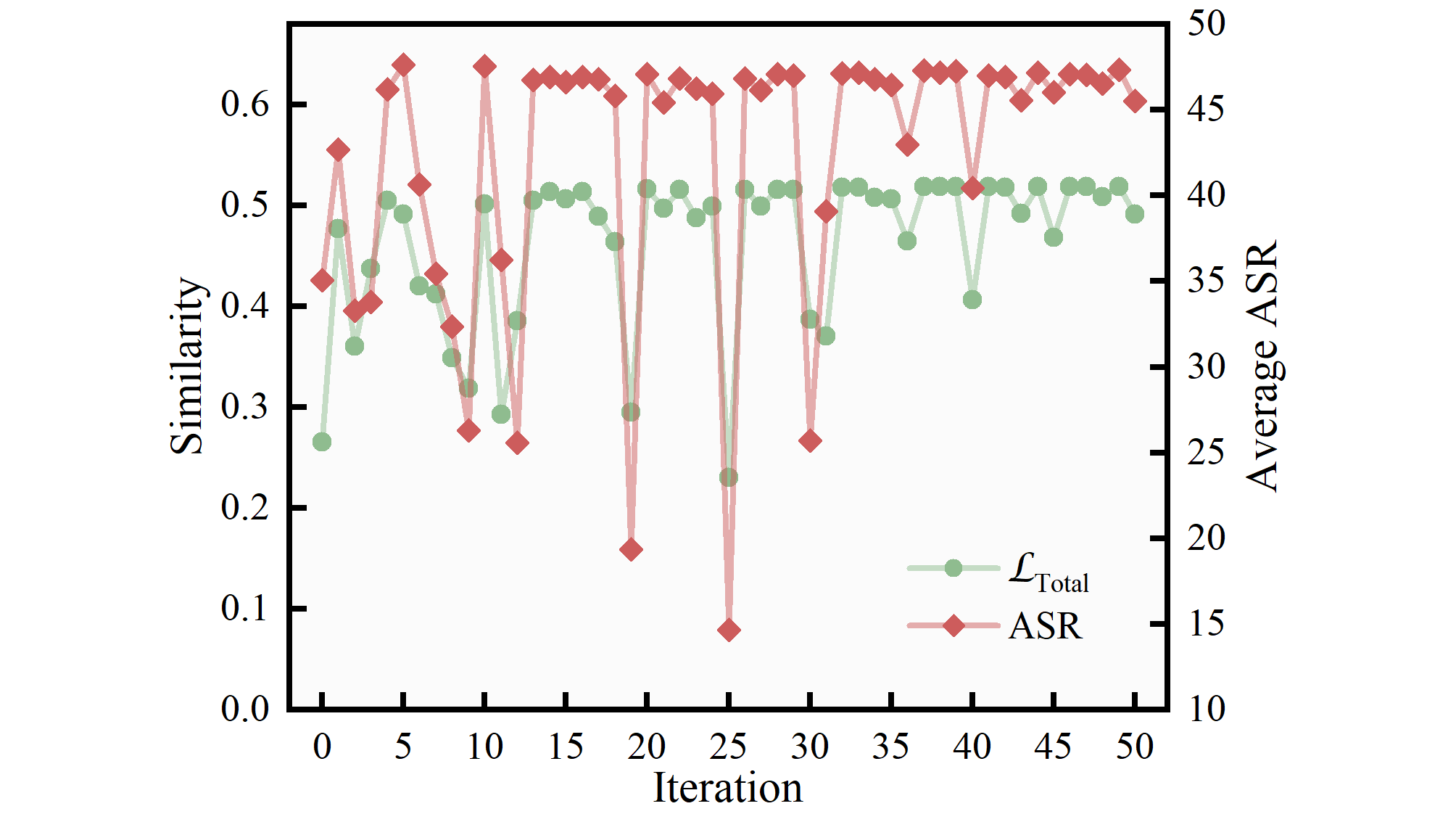}
		\caption{Average ASR (\%) and $\mathcal{L}_{\text{Data}}$ during the Bayes optimization. The optimization process involved 10 random starts and a total of 50 calls, and the initial data represents the FT-enhanced RN18.}
		\label{ASeval}
		\vspace{-4ex}
	\end{figure}

	To illustrate the effectiveness of guiding the architecture hyper-parameter search by our estimator, $\mathcal{L}_{\text{Total}}$, we show the values of average ASR and the value of $\mathcal{L}_{\text{Total}}$ during the Bayes optimization process in Fig. \ref{ASeval}.  The $\mathcal{L}_{\text{Total}}$ demonstrated the ability to mirror the trend of the average ASR and captured the fluctuation during the search. This indicates that our AS process, the $\mathcal{L}_{\text{Total}}$-guided Bayes optimization, is effective in finding well-performing architecture hyper-parameters to enhance the surrogate model's transferability.
	
	\subsection{Comparison with state-of-the-art}\label{comSOTA}	
	The comparison setting between our TEA and the state-of-the-art surrogate-side methods are reported in Table \ref{parametersetting}. A special case is IAA \cite{zhu2021rethinking} which also optimizes a transferability estimation (the alignment between gradients of data distribution and conditional density) with a similar search space as our TEA. Thus, we determined parameters for IAA and TEA by self-optimization. For the other six methods, we trained six models, ACN, SNV2, MNV2, RGN, EN, and DN121, on the synthetic set for parameter selection, and when we implemented IAA for CNX, we also set a single decay factor for all 21 blocks to avoid hard optimization.
	\begin{table}[b]
		\centering
		\caption{Parameter settings for experiments summarized in Table \ref{comparative}. Detailed information about the parameters is presented in the original papers.}
		\begin{tabular}{rccc}
			\toprule
			\makecell[c]{Method} & RN18 & RN34 & CNX \\ \hline
			SGM \cite{wu2019skip} &$\xi=0.8$ & $\xi=1.0$ & N/A \\ \cline{2-4}
			LinBP \cite{guo2020backpropagating} & layer$=$\texttt{4\_1} & layer$=$\texttt{1\_0} &  N/A \\ \cline{2-4}
			ConBP \cite{zhang2021backpropagating} & \makecell{layer$=$\texttt{4\_1}\\$\beta=3.25$} & \makecell{layer$=$\texttt{1\_0}\\$\beta=1.16$} &  N/A \\ \cline{2-4}
			IAA \cite{zhu2021rethinking} & \makecell{$\beta=32.38$\\$\bm{\xi}=[0.91,0.82,$\\$0.70, 0.31]$} & \makecell{$\beta=42.30$\\$\bm{\xi}=[1.,1.,$\\$1., 0.06]$} & \makecell{$\beta=38.71$\\$\xi=0.98$} \\ \cline{2-4}
			LRS \cite{springer2021little} & $\epsilon=0.6$ & $\epsilon=2.4$ & $\epsilon=0.4$ \\ \cline{2-4}
			DRA \cite{ZhuDRA} & $\lambda=0.1$ & $\lambda=0.05$ & $\lambda=0.05$ \\ \cline{2-4}
			DSM \cite{yang2022boosting} & $\mathcal{L}_{KL}$ & $\mathcal{L}_{KL}$ & $\mathcal{L}_{KL}$+mixup \\ \cline{2-4}
			TEA (\textbf{Ours}) & \makecell{$\beta=3.25$\\$\xi=0.75$} & \makecell{$\beta=1.16$\\$\xi=0.75$}  & \makecell{$\beta=1.04$\\$\xi=0.82$} \\
			\bottomrule
		\end{tabular}\label{parametersetting}
		\vspace{-4ex}
	\end{table}
	
	

	With the optimal parameters, the ASRs achieved by the baseline method PGD \cite{madry2018towards} are presented in Table \ref{comparative}.\footnote{All algorithms were implemented according to original papers and \textit{TransferAttackEval} \cite{zhao2023revisiting} at \url{https://github.com/ZhengyuZhao/TransferAttackEval}. The source codes of SGM, LinBP, and ConBP did not support CNX.} 
	The robustness of our target models is highlighted with the random noise, which made incorrect predictions on less than 0.2\% of the total test set. Performance corruption to RN18 (51.87\%$\rightarrow$24.97\%), RN34 (49.57\%$\rightarrow$17.63\%) and CNX (53.35\%$\rightarrow$39.02\%) from M2M to S2M is apparent based on Table \ref{comparative}, and SGM and LinBP provided minimal improvement or degradation to the baseline surrogates. Improvement to the baseline surrogate was observed for ConBP with RN34, but ConBP with RN18 resulted in degradation compared to baseline. The underlying reasons for limited improvements or degradations may be the domain shifts between the synthetic and measured data and an inability to handle the models trained on small datasets. In contrast, IAA, DRA, and DSM demonstrated effective performance improvements under the S2M setting. However, the gap between the synthetic and measured data limits the value of their impact, thereby highlighting the superiority of our TEA method. 
	
	\begin{table*}[htbp]
		\centering
		\caption{ASR (\%) against target models of architecture modification methods with PGD \cite{madry2018towards}. \underline{Underlined} data represents a white-box attack scenario and is not counted in the average. The best results are in \textbf{bold}.}
		\begin{tabular}{l|ccccccccccc|c}
			\toprule
			\multicolumn{1}{c}{} & \multicolumn{11}{c}{Victim model} & \\ \hline 
			\makecell[l]{Surrogate}  & ACN  & SNV2 & MNV2 & RGN  & EN & DN121 & RN18 & SwinT & CNX  & ViT  & VGG16  & \textbf{Avg.} \\
			\hline
			Uniform 
			&  0.00 & 0.00 & 0.00 & 0.00 & 0.22 & 0.97 & 0.00 & 0.00 & 0.07 & 0.15 & 0.00 & 0.13 \\ 
			Gaussian   & 0.00 & 0.00 & 0.00 & 0.00 & 0.22 & 0.00 & 0.00 & 0.00 & 0.00 & 0.00 & 0.00 & 0.02 \\  \hline \hline
			RN18 (M2M)  & 51.52 & 53.09 & 55.32 & 57.03 & 60.15 & 61.86 & \underline{99.63} & 28.62 & 42.01 & 60.45 & 48.70 & 51.87 \\  \hline
			RN18 (S2M)
			& 22.08 & 18.59 & 11.30 & 18.07 & 17.32 & 39.18 & 19.93 & 29.59 & 44.31 & 32.86 & 21.49 & 24.97 \\ 
			+SGM \cite{wu2019skip}  & 23.35 & 18.81 & 12.86 & 17.77 & 17.55 & 40.22 & 19.55 & 29.37 & 44.16 & 31.38 & 22.60 & 25.24 \\
			+LinBP \cite{guo2020backpropagating} & 21.49 & 16.88 & 9.96 & 15.91 & 16.06 & 39.03 & 17.99 & 27.36 & 41.41 & 30.71 & 20.82 & 23.42 \\
			+ConBP \cite{zhang2021backpropagating} &  21.34 & 16.43 & 10.19 & 14.42 & 16.06 & 39.63 & 18.51 & 27.88 & 41.34 & 31.00 & 20.22 & 23.37  \\
			+IAA \cite{zhu2021rethinking} &  30.71 & 26.39 & 14.87 & 20.07 & 27.21 & 33.61 & 19.63 & 50.04 & 55.61 & 40.52 & 29.22 & 31.63 \\
			+LRS \cite{springer2021little} & 19.78 & 20.15 & 24.39 & 30.11 & 33.23 & 44.31 & 22.23 & 27.43 & 39.55 & 42.23 & 32.04 & 30.50\\
			+DRA \cite{ZhuDRA} &  24.09 & 24.68 & 19.18 & 25.65 & 25.80 & 31.30 & 24.09 & 45.65 & 53.09 & 46.10 & 33.75 & 32.13 \\
			+DSM \cite{yang2022boosting} &  31.52 & 20.30 & 12.34 & 23.79 & 24.76 & 36.80 & 22.60 & 46.32 & 51.82 & 46.69 & 26.91 & 31.26 \\
			\cellcolor[rgb]{0.9,0.9,0.9}+TEA  (\textbf{Ours})   &\cellcolor[rgb]{0.9,0.9,0.9}\textbf{44.46} &\cellcolor[rgb]{0.9,0.9,0.9}\textbf{29.74} &\cellcolor[rgb]{0.9,0.9,0.9}\textbf{29.22} &\cellcolor[rgb]{0.9,0.9,0.9}\textbf{43.12} &\cellcolor[rgb]{0.9,0.9,0.9}\textbf{54.50} &\cellcolor[rgb]{0.9,0.9,0.9}\textbf{41.34} &\cellcolor[rgb]{0.9,0.9,0.9}\textbf{35.61} &\cellcolor[rgb]{0.9,0.9,0.9}\textbf{64.24} &\cellcolor[rgb]{0.9,0.9,0.9}\textbf{67.58} &\cellcolor[rgb]{0.9,0.9,0.9}\textbf{62.01} &\cellcolor[rgb]{0.9,0.9,0.9}\textbf{49.14} &\cellcolor[rgb]{0.9,0.9,0.9}\textbf{47.36}  \\ \hline 
			\hline 
			
			RN34 (M2M) &  40.22 & 49.07 & 48.62 & 56.28 & 52.64 & 52.79 & 55.32 & 30.78 & 48.33 & 62.83 & 48.40 & 49.57  \\  \hline
			RN34 (S2M)
			&  8.62 & 6.02 & 7.58 & 6.10 & 12.49 & 39.26 & 7.14 & 22.16 & 42.68 & 24.61 & 17.25 & 17.63 \\ 
			+SGM \cite{wu2019skip} & 8.10 & 6.02 & 7.14 & 6.39 & 12.27 & 39.03 & 6.62 & 22.16 & 43.12 & 23.57 & 17.92 & 17.49 \\
			+LinBP \cite{guo2020backpropagating} & 8.92 & 5.65 & 5.43 & 7.14 & 10.11 & 37.47 & 7.06 & 27.51 & 31.08 & 14.94 & 15.09 & 15.49 \\
			+ConBP \cite{zhang2021backpropagating} &  24.01 & 22.53 & 17.70 & 24.54 & 31.38 & \textbf{43.20} & 29.96 & 56.65 & 61.04 & 37.77 & 32.19 & 34.63 \\
			+IAA \cite{zhu2021rethinking} &  28.25 & 20.97 & 15.39 & 25.72 & 27.29 & 36.51 & 24.61 & 41.78 & 48.85 & 42.90 & 36.13 & 31.67 \\
			+LRS \cite{springer2021little} & 14.57 & 18.07 & 21.78 & 33.75 & 38.88 & 43.49 & 13.68 & 24.16 & 38.14 & 51.60 & 25.72 & 29.44 \\
			+DRA \cite{ZhuDRA} &26.02 & 13.09 & 11.67 & 16.21 & 22.60 & 29.14 & 15.61 & 31.60 & 44.76 & 39.33 & 31.00 & 25.55 \\
			+DSM \cite{yang2022boosting} &  25.65 & 22.97 & 11.52 & 15.76 & 24.76 & 38.74 & 22.83 & 37.84 & 47.66 & 37.84 & 32.71 & 28.94 \\
			\cellcolor[rgb]{0.9,0.9,0.9}+TEA (\textbf{Ours})  &\cellcolor[rgb]{0.9,0.9,0.9}\textbf{38.44} &\cellcolor[rgb]{0.9,0.9,0.9}\textbf{27.73} &\cellcolor[rgb]{0.9,0.9,0.9}\textbf{36.06} &\cellcolor[rgb]{0.9,0.9,0.9}\textbf{36.43} &\cellcolor[rgb]{0.9,0.9,0.9}\textbf{42.23} &\cellcolor[rgb]{0.9,0.9,0.9}42.08 &\cellcolor[rgb]{0.9,0.9,0.9}\textbf{32.71} &\cellcolor[rgb]{0.9,0.9,0.9}\textbf{58.96} &\cellcolor[rgb]{0.9,0.9,0.9}\textbf{66.25} &\cellcolor[rgb]{0.9,0.9,0.9}\textbf{77.99} &\cellcolor[rgb]{0.9,0.9,0.9}\textbf{48.18} &\cellcolor[rgb]{0.9,0.9,0.9}\textbf{46.10}  \\ \hline 
			\hline 
			
			CNX (M2M) &  30.63 & 47.96 & 49.59 & 49.59 & 79.33 & 42.45 & 33.09 & 100.00 & \underline{99.85} & 59.03 & 41.86 & 53.35  \\  \hline
			CNX (S2M)
			& 25.20 & 24.24 & 17.17 & 22.30 & 35.54 & 27.36 & 14.57 & 81.04 & 81.56 & 58.29 & 41.93 & 39.02 \\ 
			+IAA \cite{zhu2021rethinking} &  27.14 & 25.06 & 20.00 & 24.09 & 37.17 & 27.06 & 15.99 & 82.01 & 83.57 & 59.63 & 42.45 & 40.38 \\
			+LRS \cite{springer2021little} & 29.74 & 28.10 & 31.23 & 35.69 & 42.23 & 39.26 & 22.53 & 74.87 & 75.69 & 72.64 & 38.51 & 44.59  \\
			+DRA \cite{ZhuDRA} & 31.38 & 27.29 & 26.17 & 28.85 & 34.50 & \textbf{30.78} & 17.77 & \textbf{90.33} & 85.06 & 61.26 & 37.92 & 42.85 \\
			+DSM \cite{yang2022boosting} &  30.48 & 28.10 & 20.00 & 25.35 & 37.84 & 30.19 & 17.84 & 83.57 & 82.16 & 63.35 & 43.05 & 41.99 \\
			\cellcolor[rgb]{0.9,0.9,0.9}+TEA (\textbf{Ours})   &\cellcolor[rgb]{0.9,0.9,0.9}\textbf{36.58} &\cellcolor[rgb]{0.9,0.9,0.9}\textbf{41.34} &\cellcolor[rgb]{0.9,0.9,0.9}\textbf{44.09} &\cellcolor[rgb]{0.9,0.9,0.9}\textbf{54.94} &\cellcolor[rgb]{0.9,0.9,0.9}\textbf{49.81} &\cellcolor[rgb]{0.9,0.9,0.9}26.39 &\cellcolor[rgb]{0.9,0.9,0.9}\textbf{24.39} &\cellcolor[rgb]{0.9,0.9,0.9}86.84 &\cellcolor[rgb]{0.9,0.9,0.9}\textbf{89.37} &\cellcolor[rgb]{0.9,0.9,0.9}\textbf{79.93} &\cellcolor[rgb]{0.9,0.9,0.9}\textbf{46.25} &\cellcolor[rgb]{0.9,0.9,0.9}\textbf{52.72} \\				
			\bottomrule
		\end{tabular}\label{comparative}
	\end{table*}
	
	Overall, the TEA attack outperformed the state-of-the-art architecture modification competitors and \textit{significantly} improved the S2M transferability, boosting the baseline average ASRs of 24.97\% (RN18), 17.63\% (RN34), and 39.02\% (CNX) to 47.36\%, 46.10\%, and 52.72\%, respectively. It is also important to note that the improved results are near the baseline performance of the M2M scenario. Moreover, the TEA generalized well within and across the ResNet family, even when there was a significant gap in model structure design.
	
	\begin{figure*}[tbp]
		\centering
		\includegraphics[width=1\linewidth]{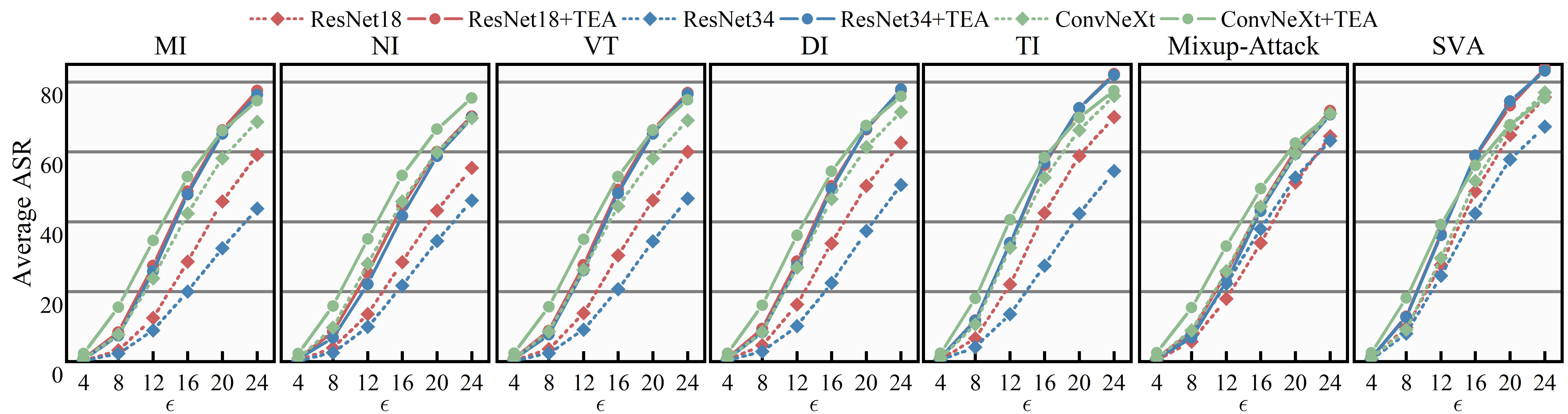}
		\caption{The \textit{Average ASR (\%) vs. perturbation budget} (pixel values of $\epsilon/255$) curves resulting from combinations of the competitors with our enhanced surrogate models. }
		\label{variousbudgets}
	\end{figure*}
	\subsection{Compatibility with other attacks}\label{compare}
\begin{figure}[tbp]
		\centering
		\includegraphics[width=1\linewidth]{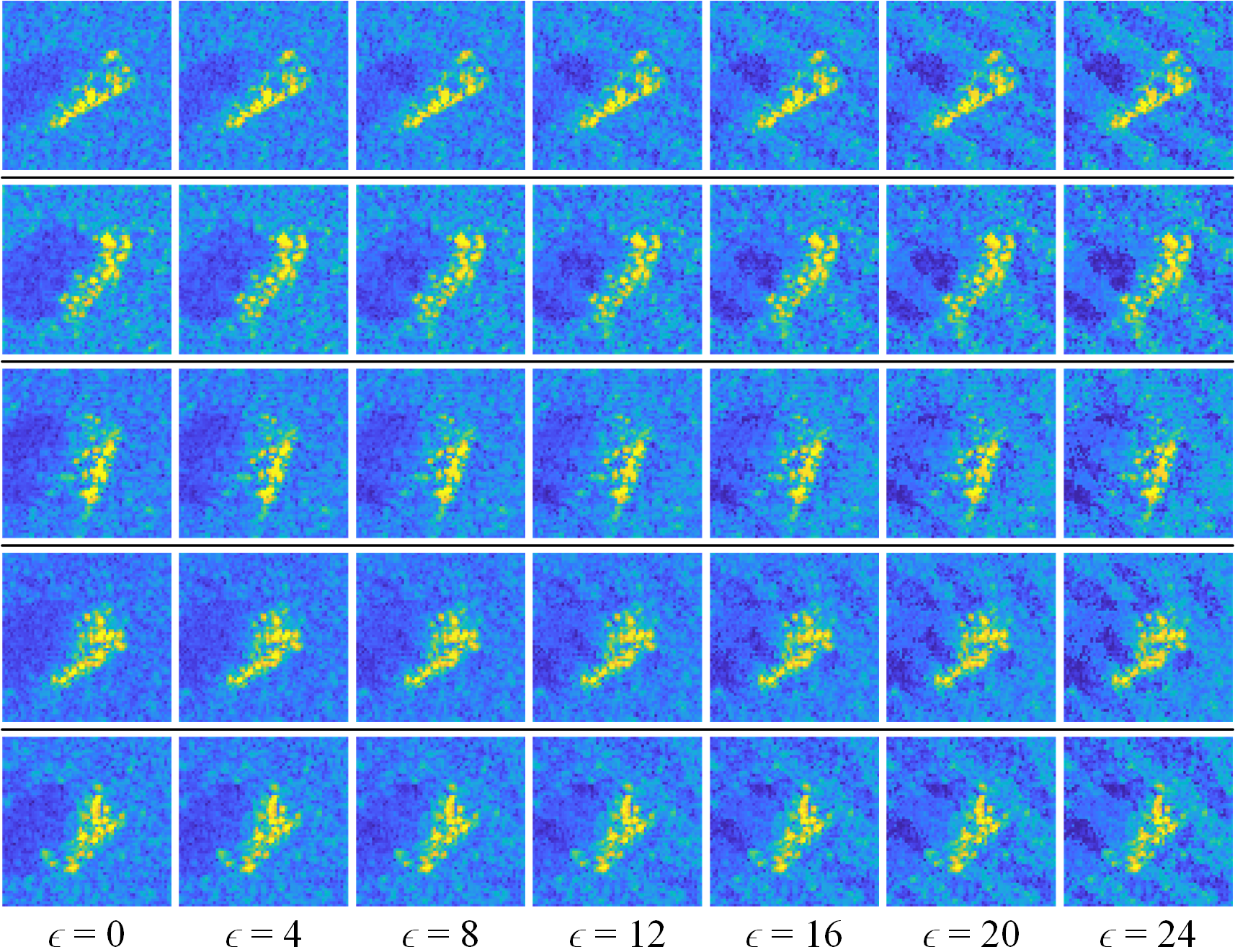}
		\caption{Adversarial examples generated by RN18+TEA and PGD with various perturbation budgets. Note that $\epsilon=0$ represents the clean images.}
		\label{adversarialexamples}
	\end{figure}

	This section studies the compatibility of TEA with other advanced gradient-based attacks in the computer vision and remote sensing communities, including MI \cite{momentum2018dong}, NI \cite{nestrov2019lin}, VT \cite{Wang_2021_CVPR}, DI \cite{inputdiversity2019xie}, TI \cite{evading2019dong}, Mixup-Attack \cite{rsaa2}, and speckle-variant attack (SVA) \cite{sva}. We show the average ASR of these attacks with different perturbation budgets in Fig. \ref{variousbudgets}, and the results suggest that with all three studied surrogate models, the seven advanced attacks can benefit from our TEA method, demonstrating the compatibility of our method and its ability to create more powerful attacks. The TEA-enhanced surrogates enable us to achieve comparable results with smaller perturbations, and Fig. \ref{adversarialexamples} showcases the adversarial examples generated under different perturbation budgets. The images demonstrate that the perturbation budget plays a crucial role in stealthiness, and our method helped to balance the attack capability and stealthiness in the S2M setting. For instance, the average ASR of 58.81\% achieved by TI attack based on RN34+TEA under $\epsilon=16/255$ was higher than the 54.52\% average ASR of original RN34 under $\epsilon=24/255$.
	
\begin{figure}[tbp]
		\centering
		\includegraphics[width=0.95\linewidth]{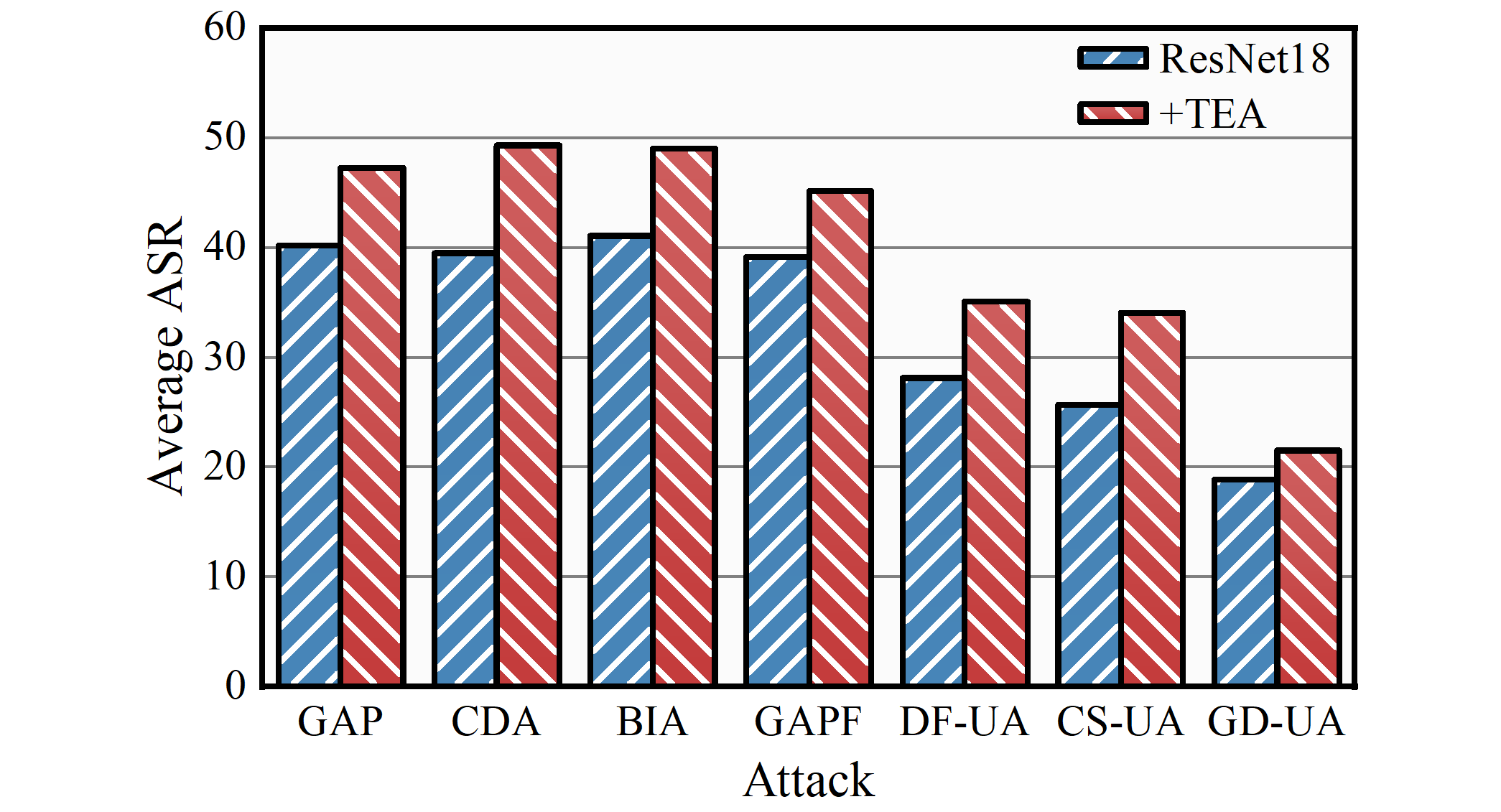}
		\caption{Average ASR  (\%) against target models of the generative and universal attacks with the S2M setting and RN18 as the surrogate model.}
		\label{capatibilitywithotherattacks}
	\end{figure}

	Although we designed our approach for gradient-based attacks, we also investigated whether the approach was compatible with other categories of attacks. Fig. \ref{capatibilitywithotherattacks}\footnote{All four generative attacks were implemented with the same generator architecture at \url{https://github.com/Muzammal-Naseer/CDA/blob/master/generators.py} and the same initialization for 20 epochs training with the Adam optimizer and a learning rate of 0.001. The features extracted at layer 4 were targeted for BIA and GAFP to attack. The universal perturbations were optimized over 20 epochs with the Adam optimizer and a learning rate of 0.01.} shows the attack performance of our surrogates equipped with four generative attacks (generative adversarial perturbations (GAP) \cite{poursaeed2018generative}, CDA \cite{CDA2019}, beyond ImageNet attack (BIA) \cite{zhang2021beyond}, and generative adversarial feature perturbations (GAFP) \cite{LTAP2021}) and three universal attacks (dominant feature attack (DF-UA) \cite{domainfeatureuap}, cosine similarity attack (CS-UA) \cite{datafreeuap}, and generalizable data-free attack (GD-UA) \cite{GDUAP}). The performance improvements enabled by TEA are apparent for all methods in Fig. \ref{capatibilitywithotherattacks}, where TEA enabled an improvement of 7.98\% for BIA and 8.34\% for CS-UA. The best average ASR of 49.28\% was achieved by CDA with RN18+TEA, and the highest improvement was 9.78\% to CDA. Note that there was a mismatch in attack objectives, as BIA and GAFP primarily manipulate features triggered at intermediate layers rather than gradients. Nonetheless, these findings demonstrate the compatibility of TEA with generative and universal attacks and highlight its potential to enhance model transferability at feature and gradient levels. Note that it is also reasonable to expect that further improvements for these attacks with TEA could be unlocked with specialized adaptations on the same attack objective. Meanwhile, the gain of TEA to universal attacks can also facilitate the more challenging unpaired transfer scenarios where the attacker does not know the type of objects the victim model is trained on \cite{peng2022empirical}.
	
	\subsection{Parameter sensitivity}\label{sensitivity}
	The rationality and stability of the parameter selection strategy are crucial for TEA optimization since the target data and models are inaccessible. We did not obtain the optimal surrogate; instead, we reported the results given by our strategy. In this section, we report the parameter sensitivity analysis of RN18 of our TEA parameter selection strategy. Note that all results reported herein were achieved by PGD \cite{madry2018towards} attack. 
	
	\begin{figure}[tbp]
		\centering	
		\includegraphics[width=1\linewidth]{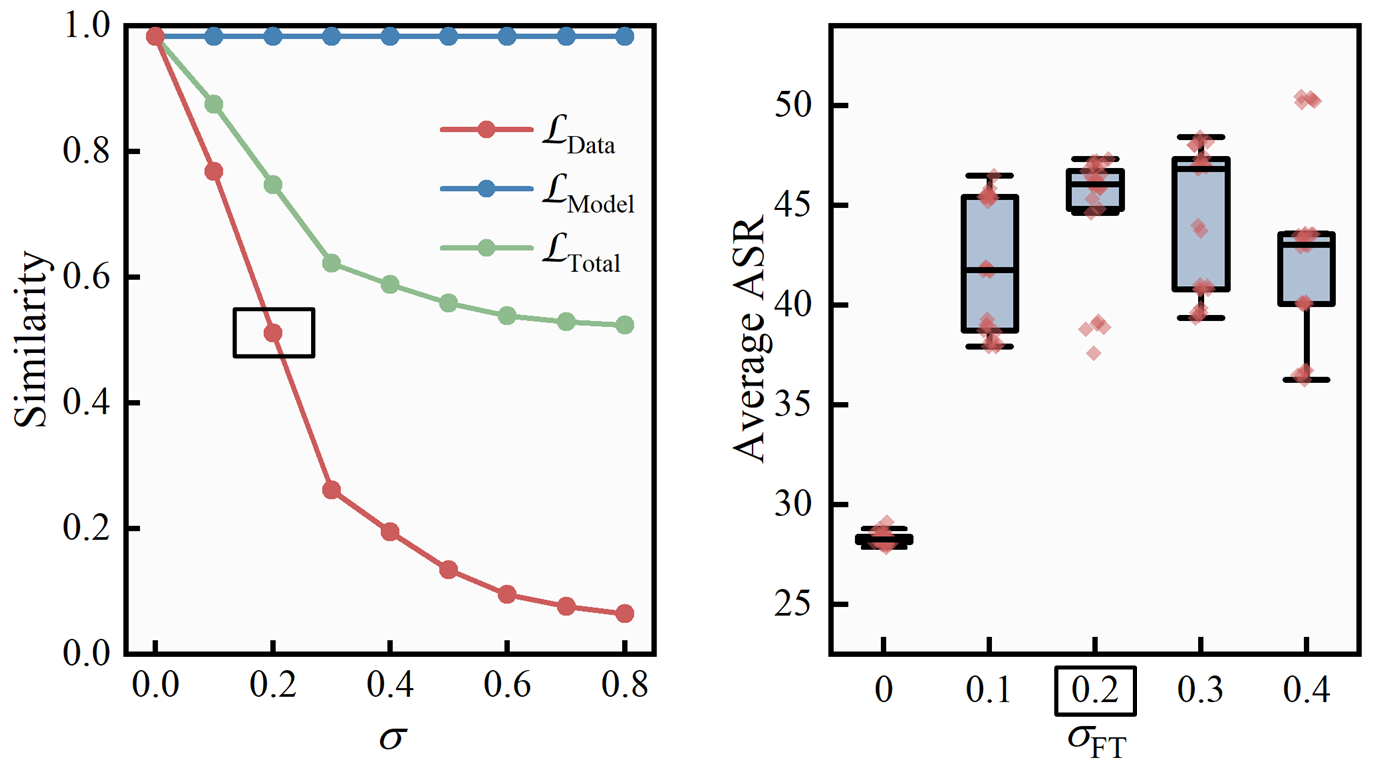}
		\caption{Stability study for the FT process: (\textbf{Left}) the values of $\mathcal{L}_{\text{Data}}$ and $\mathcal{L}_{\text{Model}}$ tested with the synthetic dataset-trained RN18 on substitute data with a standard deviation of $\sigma$ and (\textbf{Right}) box-plots of average ASR (\%) against eleven target models with different $\sigma_{\text{FT}}$ in fine-tuning. \boxed{\text{Boxed}} data indicates our choice. Symbol clutters in the box-plot were the result of different random Bayes optimization trails for the same surrogate.}
		\label{FTtrain}
	\end{figure}
	
	\subsubsection{Selection of $\sigma_{\text{FT}}$}
	Recall that we select $\sigma_{\text{FT}}$ such that $\mathcal{L}_{\text{Data}}(\sigma_{\text{FT}}) \approx 0.5$. \textcolor{black}{To achieve this, we performed a simple search that calculated the proposed estimators when changing $\sigma$, as shown in the left sub-figure of Fig. \ref{FTtrain}, and $\sigma_{\text{FT}}=0.2$ satisfied our strategy. Note that $\mathcal{L}_{\text{Model}}=1$ here because $f^{\ast\text{sur}}$ and $f^{\text{sur}}$ were the same at this time (before optimization).} To investigate the effectiveness of our choice, we repeated the FT process for $\sigma_{\text{FT}}=\{0, 0.1, 0.2, 0.3, 0.4\}$ with $\lambda=1$ five times with different seedings. We then completed the AS stage with $\sigma_{\text{AS}}=0.3$ applied to the 25 obtained surrogate models with five different seeds to investigate the influence of randomness in Bayes optimization. Results show that the randomness in FT plays an important role in influencing the final results, but in most cases, five repeated Bayes optimization procedures achieved similar results, as indicated by clusters in the box plot. One can find that although our choice did not achieve the best result, it was effective and more stable than other choices.

	\begin{figure}[tbp]
		\centering	
		\includegraphics[width=1\linewidth]{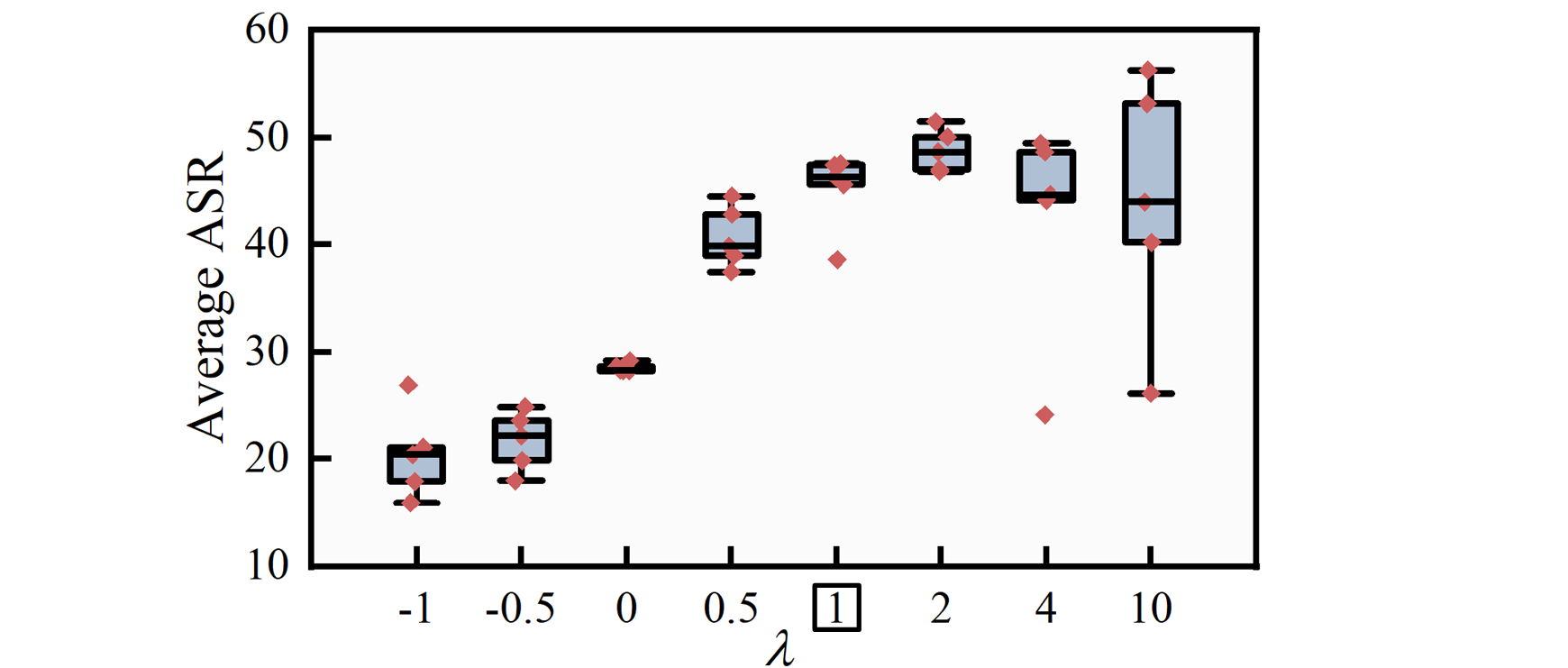}
		\caption{Box-plot of average ASR (\%) resulted from different $\lambda$ in fine-tuning. \boxed{\text{Boxed}} data indicates our choice.}
		\label{lambda}
	\end{figure}
	
	\subsubsection{Selection of $\lambda$}
	\textcolor{black}{To investigate the influence of $\lambda$, we ran five random FTs for $\lambda=\{-1, -0.5, 0, 0.5, 1, 2, 4, 10\}$ with $\sigma_{\text{FT}}=0.2$ and performed a single AS for each since the AS randomness had less of an impact on the results.} The results, depicted in Fig. \ref{lambda}, show performance degradation for $\lambda<0$ and performance improvements for $\lambda>0$. This proves the effectiveness of our loss design of $\mathcal{L}_{\text{FT}}$, and although our choice of $\lambda=1$ did not obtain the best result (56.31\% at $\lambda=10$), it exhibited moderate effectiveness and the best overall stability, indicating a satisfactory choice for blind optimization.
	
	\subsubsection{Selection of $\sigma_{\text{AS}}$}
\begin{figure}
		\centering	
		\includegraphics[width=1\linewidth]{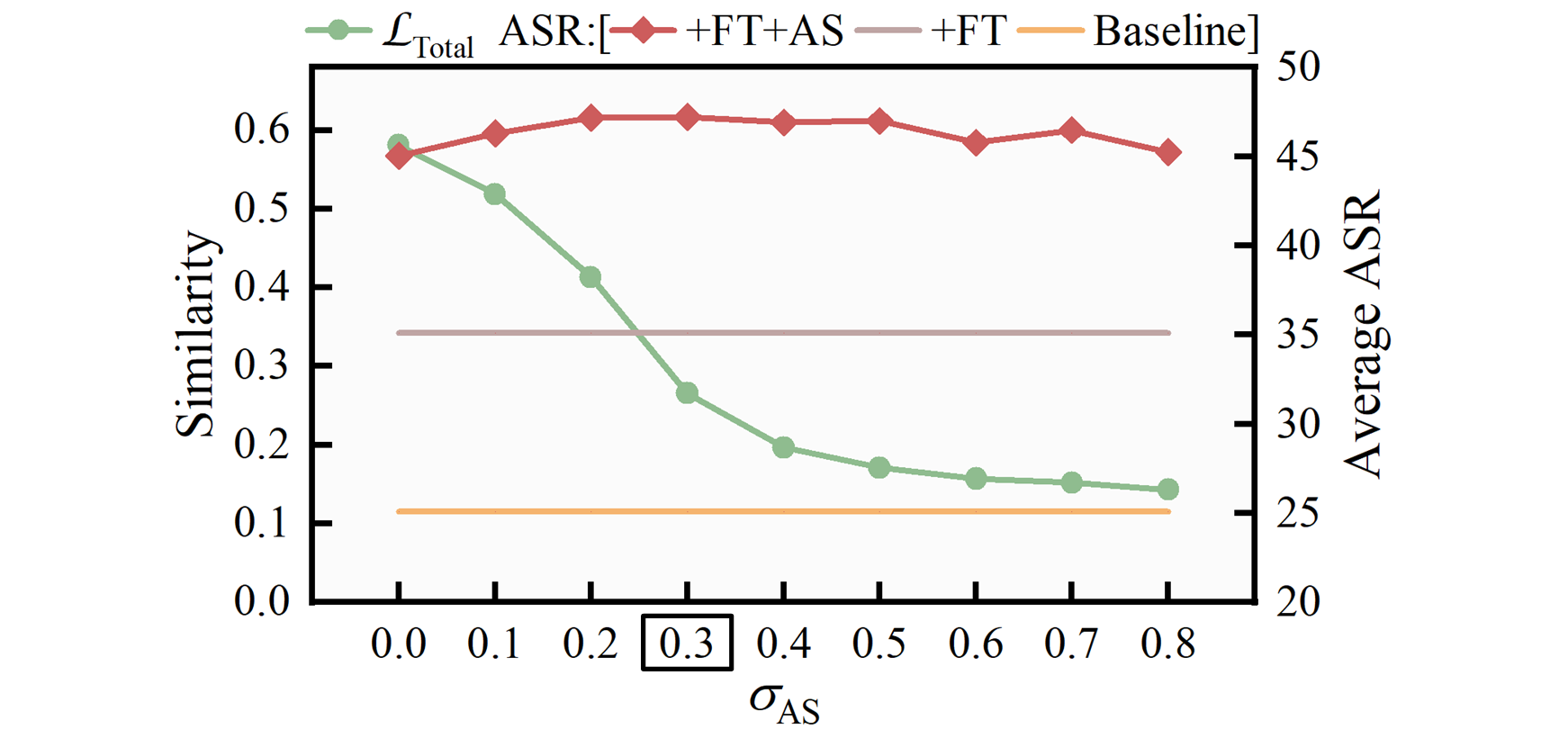}
		\caption{Average ASR (\%) as function of $\sigma_{\text{AS}}$ in the AS stage. \boxed{\text{Boxed}} data indicates our choice.}
		\label{ASsensi}
	\end{figure}

	To investigate the effect of $\sigma_{\text{AS}}$ in the AS process, we selected a model that was trained with $\sigma_{\text{FT}}=0.2$ and $\lambda=1$ and performed AS with various $\sigma_{\text{AS}}$ values. \textcolor{black}{Using a similar strategy as the selection for $\sigma_{\text{FT}}$, we also made our selection for $\sigma_{\text{AS}}$ where $\mathcal{L}_{\text{Total}}\approx0.2$ according to our strategy.}
	The resulting average ASR is shown in Fig. \ref{ASsensi}, where the green curve represents the value of $\mathcal{L}_{\text{Total}}$. The AS optimization was barely impacted by $\mathcal{L}_{\text{Model}}$ with $\sigma_{\text{AS}}=0$, and all nine results ranged from average ASRs of 45.05\% to 47.21\% with the best at $\sigma_{\text{AS}}=0.3$ and $\mathcal{L}_{\text{Total}}=0.2654$. Therefore, $\sigma_{\text{AS}}$ did not have a significant effect on the results, and these results align with our earlier analysis that a too large or too small initial $\sigma_{\text{AS}}$ will hinder the optimization due to a narrow window to find a better solution or the saturation phenomenon. Selecting an initial $\sigma_{\text{AS}}$ in the range of 0.2 to 0.5 is the most suitable choice for finding good architecture hyper-parameters.

	\subsection{Ablation study}
	Since the effects of the TEA components and parameters have been investigated in Sections \ref{evaluation} and \ref{sensitivity}, we performed a component-level ablation study to identify how each of the components of TEA (\textit{i.e.}, the estimator, FT, and AS) affected the final performance. Table \ref{ablation} reports the average ASR of PGD, DI, and TI in four cases, where each corresponds to a combination of the components of TEA. 
	From the table, we conclude that the FT and AS are effective and provide considerable improvements when working together. The improvements enabled by $\mathcal{L}_{\text{Model}}$ to FT+$\mathcal{L}_{\text{Data}}$ are also clearly demonstrated in the table, which verifies the effectiveness of our design on the estimator $\mathcal{L}_{\text{Total}}$. Note that $\mathcal{L}_{\text{Data}}$ alone failed to boost the transferability of DI and TI with CNX.
	
	\begin{table}[htbp]
		\centering
		\caption{Effects of the components of TEA. Best results are in \textbf{bold}.}
		\label{ablation}
		\begin{tabular}{c|ccc|ccc}
			\toprule		
			\multicolumn{2}{c}{} & \multicolumn{2}{c}{AS} &  & &   \\ \cmidrule(r){3-4}
			Surr. & FT & $\mathcal{L}_{\text{Data}}$ & $\mathcal{L}_{\text{Model}}$ & PGD  &  DI & TI \\	\midrule
			\multirow{5}*{RN18} & & & & 24.97 &  33.77 & 42.50 \\ 
			& \Checkmark & & & 35.07  &  39.33 & 44.14  \\ 
			& \Checkmark & \Checkmark & & 38.75  &  40.55 & 46.96 \\
			& \Checkmark & \Checkmark &\Checkmark & \textbf{47.36}  &  \textbf{50.18} & \textbf{56.23} \\ \hline \hline  
			\multirow{5}*{RN34} & & & &17.63  &  22.54 & 27.41  \\ 
			& \Checkmark & & & 31.19  &  37.89 & 39.87  \\ 
			& \Checkmark & \Checkmark & & 43.06  &  46.52 & 54.11 \\
			& \Checkmark & \Checkmark &\Checkmark & \textbf{46.10}  & \textbf{49.37} & \textbf{56.81}  \\ \hline \hline  
			\multirow{5}*{CNX} & & & & 39.02  &  46.48 & 52.61  \\ 
			& \Checkmark & & & 46.62  &  51.58 & 57.82  \\ 
			& \Checkmark & \Checkmark & & 47.94 & 49.02  & 54.07 \\
			& \Checkmark & \Checkmark &\Checkmark & \textbf{52.72}  &  \textbf{54.42} & \textbf{58.34} \\
			\bottomrule
		\end{tabular}
	\end{table}%
	\section{Discussion}\label{discussion}
	
	In this section, we provide additional clarity and discussion on the TEA with the up-to-date adversarial transferability theory, the relationship between generalization and transferability in the S2M setting, and the physical applicability of this study.
	\subsection{Understanding the effectiveness of the TEA}\label{understanding}
	\subsubsection{Model smoothness and gradient similarity}
	Here, we utilize the latest theoretical understanding of adversarial transferability to analyze how the proposed TEA boosts the S2M transferability. The model smoothness and gradient similarity, as defined in Eq. \eqref{eq2}, are positively correlated to the lower bound of adversarial transferability \cite{zhang2024does,yang2021trs}. It has been shown that model smoothness in input and weight space is highly complementary in prompting transferability \cite{zhang2024does}, but the intangible nature of gradient similarity towards an unknown target model still makes it difficult to obtain a better surrogate. In this paper, we show that the gradient similarity can be implicitly transferred to the input and weight space smoothness by the TEA in the S2M setting, and the gradient similarity towards the measured data-trained target model is disentangled to data and model discrepancies. The data and model discrepancies can then be measured by $\mathcal{L}_{\text{Data}}$ and $\mathcal{L}_{\text{Model}}$, respectively, and optimized over the architecture hyper-parameter search space. 
	\begin{figure}[tbp]
		\centering
		\subfloat[The \textit{loss vs. weight variation} landscapes over the (\textbf{Left}) synthetic and (\textbf{Right}) measured datasets. Note that we randomly sampled the direction for weight variation \cite{losslandscape}.]{\includegraphics[width=0.85\linewidth]{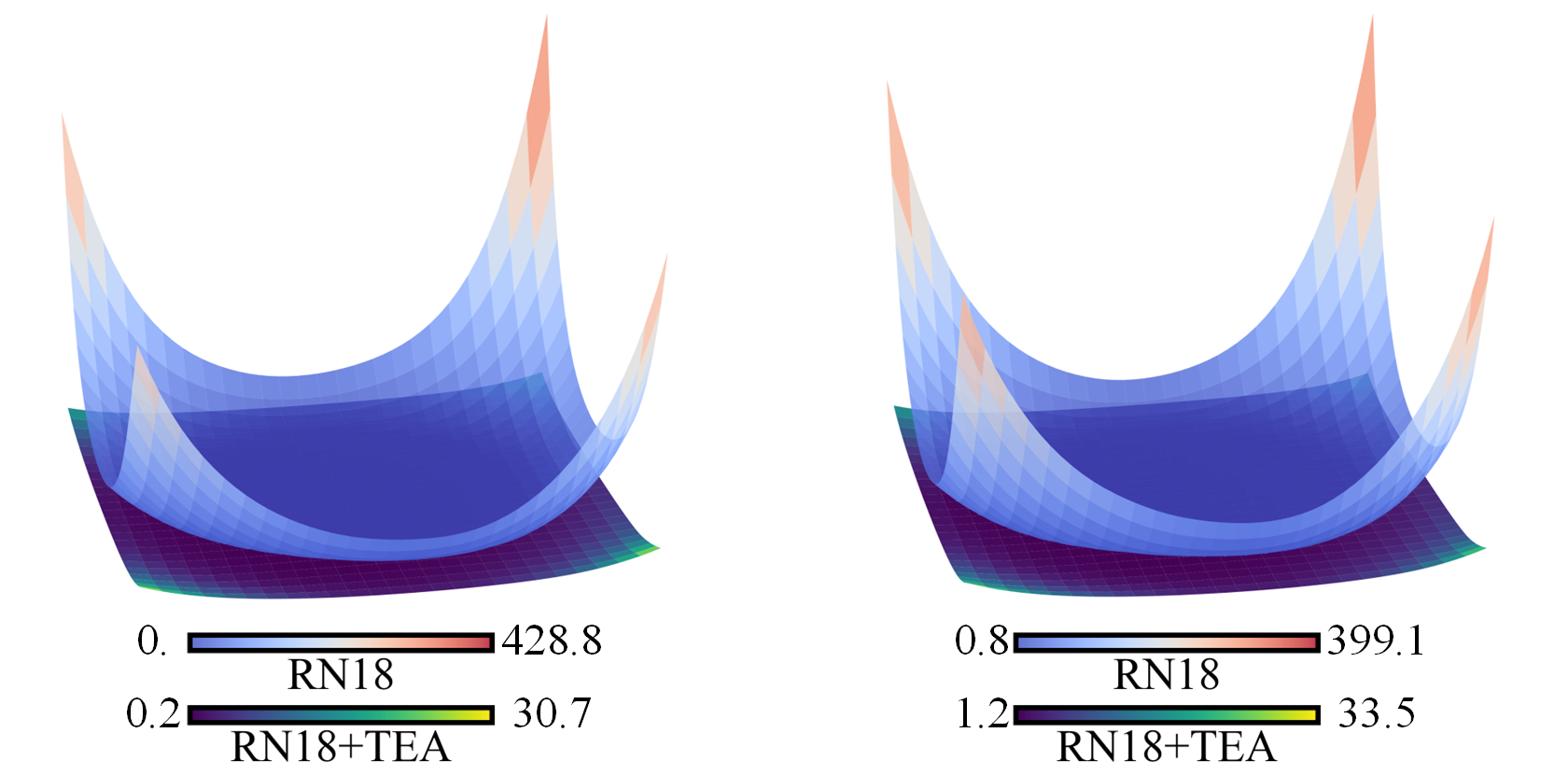}}\\
		
		\subfloat[The \textit{loss vs. input variation} landscapes over the (\textbf{Left}) synthetic and (\textbf{Right}) measured datasets. We sampled the adversarial direction and its orthogonal direction to calculate the loss values.]{\includegraphics[width=0.85\linewidth]{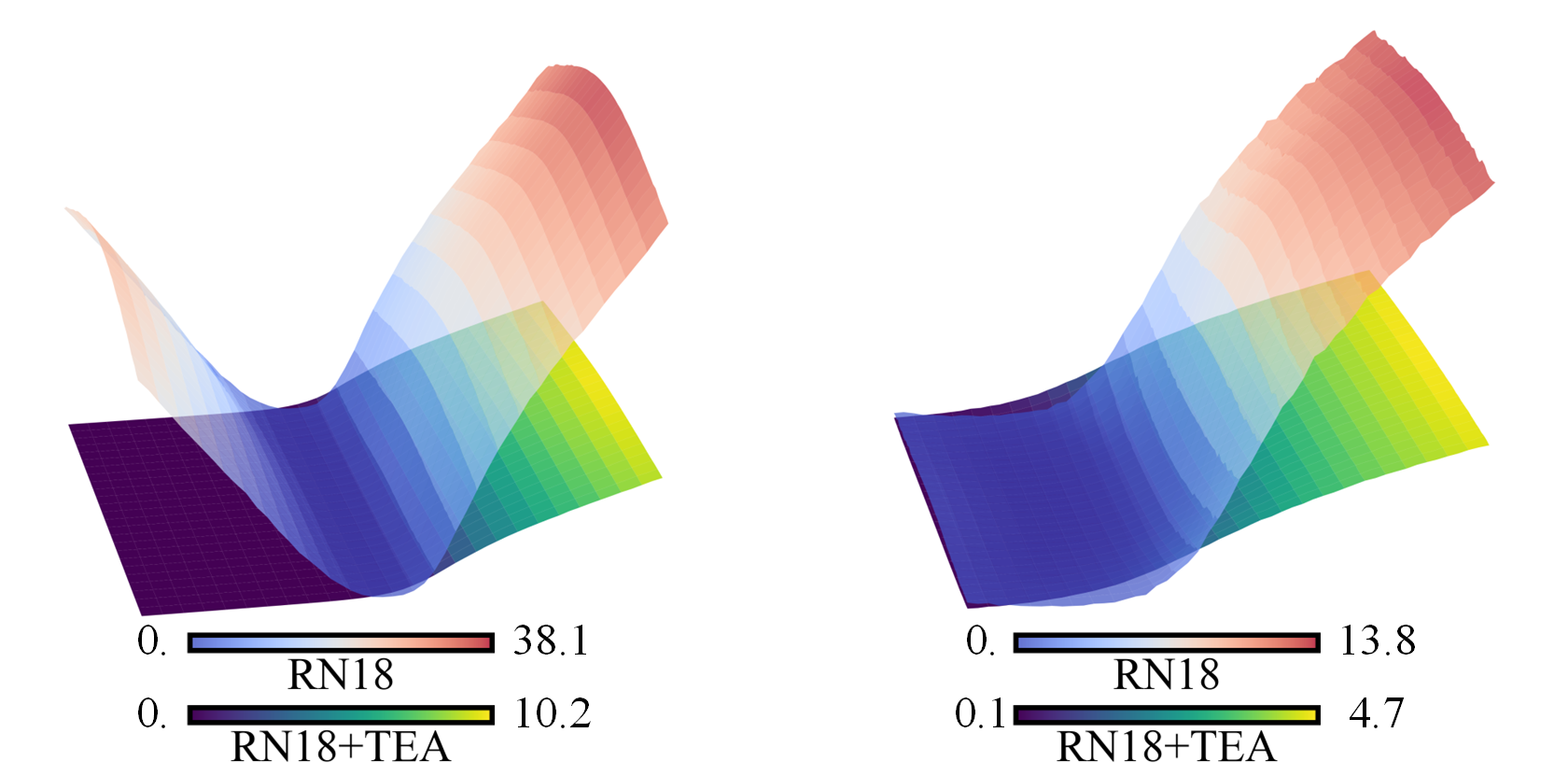}}
		\caption{Loss landscapes of the original and TEA-enhanced RN18 over the synthetic and measured datasets.}
		\label{landscapes}
	\end{figure}
	\begin{figure}[tbp]
		\centering
		\includegraphics[width=0.9\linewidth]{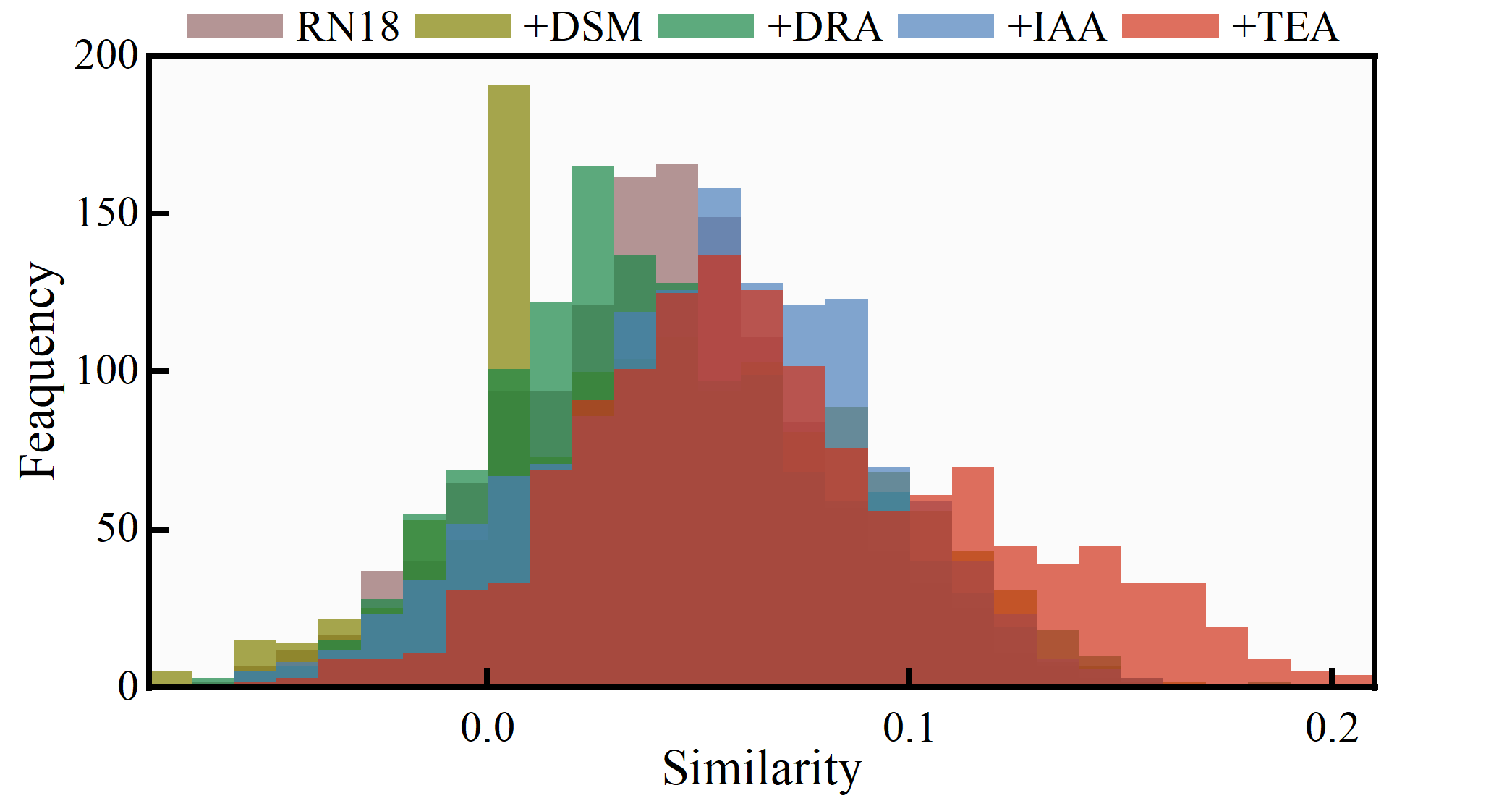}
		\caption{Frequency histogram of the average gradient similarity for eleven target models.}
		\label{histogram}
	\end{figure}
	Here, we reconsider our estimator from the model smoothness perspective. Given fixed model weights, $\mathcal{L}_{\text{Data}}$ restricts the variation in gradient directions w.r.t. the original $\bm{x}^\text{syn}$ and randomly sampled neighbors of $\bm{x}^\text{sub}$, improving the input space smoothness. Furthermore, for given input data, $\mathcal{L}_{\text{Model}}$ restricts variation in gradient directions w.r.t. the fine-tuned surrogate model, $f^{\text{sur}}_{\bm{\Theta}^\ast, \bm{\Lambda}}$, and its enhanced version, $f^{\text{sur}}_{\bm{\Theta}^\ast, \bm{\Lambda}^\ast}$, promoting weight (architecture hyper-parameters) space smoothness. This analysis is relatively intuitive, and we give empirical evidence in Fig. \ref{landscapes}. The TEA-enhanced RN18 is significantly smoother than the original model in the weight space and the input space, and this manifests as improvements in gradient alignment, as indicated in Fig. \ref{histogram}.
	
	\begin{figure}[tbp]
		\centering
		\includegraphics[width=1\linewidth]{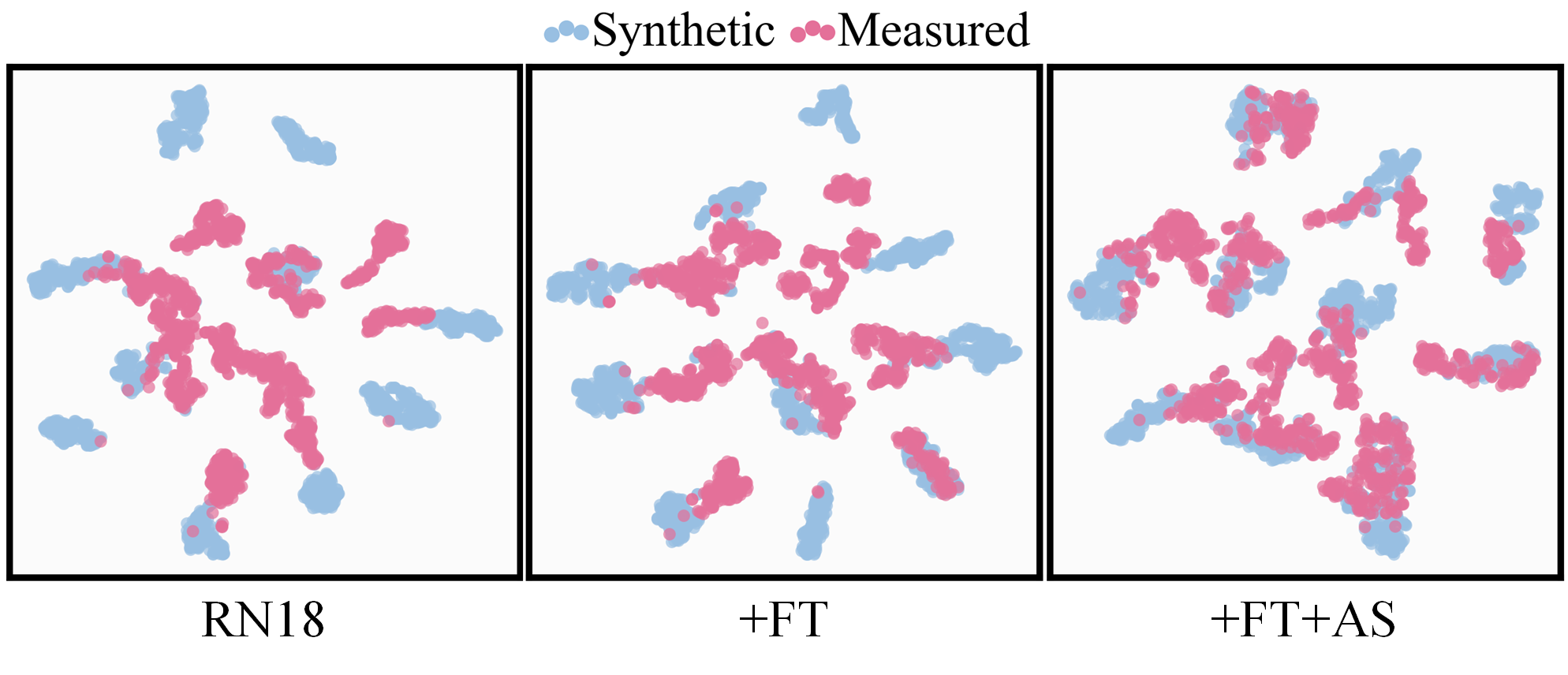}
		\caption{Feature embedding visualized by t-SNE \cite{van2008visualizing}.}
		\label{tsne}
	\end{figure}
	\begin{figure}[tbp]
		\centering
		\includegraphics[width=1\linewidth]{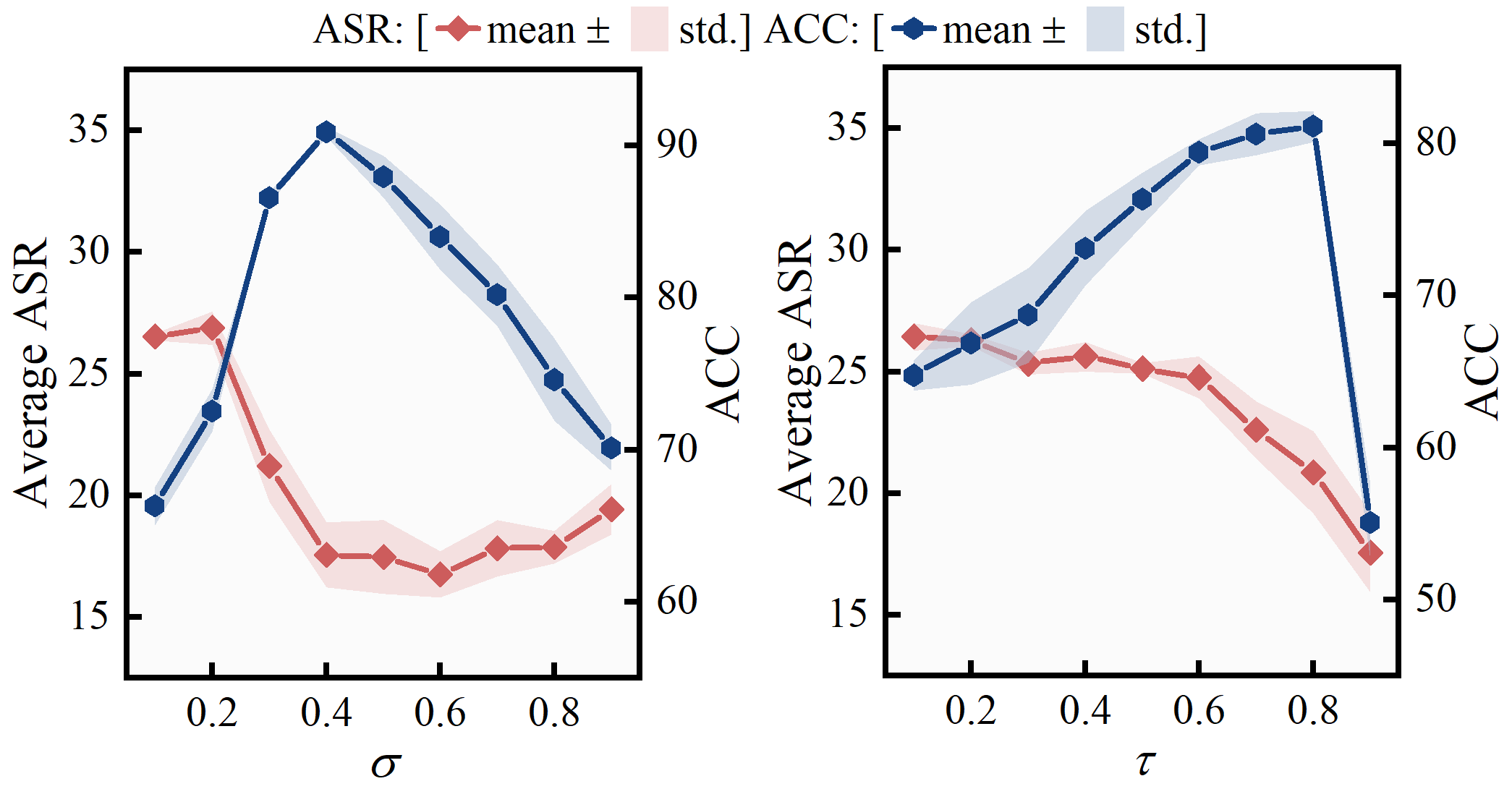}
		\caption{Average ASR and ACC both in \% of (\textbf{Left}) Gaussian noise augmentation with std. $\sigma$ and (\textbf{Right}) adding dropout layers with drop rate $\tau$. Results are averaged over 5 trials.}
		\label{guasaug_dropout}
	\end{figure}
	
	\subsubsection{t-distributed stochastic neighbor embedding visualization}
	Here, we visualize the feature embedding of the synthetic and measured data given by the original surrogate and its TEA-enhanced variation using t-distributed stochastic neighbor embedding (t-SNE) \cite{van2008visualizing}. Specifically, we fed both synthetic and measured datasets to the original, +FT, and +FT+AS models and visualized the feature embedding output by the penultimate layers. In this approach, higher degrees of fusion between the two distributions indicates better generalization ability from the synthetic to the measured domain of the model. As shown in Fig. \ref{tsne}, the original synthetic dataset-trained RN18 surrogate yielded clearly distinct feature embedding of the synthetic and measured data, and the FT and AS processes were able to enhance the generalization, resulting in more similar embedding of the synthetic and measured paired data. This advantage is beneficial to S2M evaluation as well as the SAMPLE recognition challenge.
	
	\subsection{Generalization \textit{vs.} transferability}\label{genvstrans}
	Recall that the substitute data, noised copy of synthetic data, is utilized in fine-tuning the surrogate and measuring S2M transferability. It is valid to ask if exploiting the gradient direction in our design is necessary, as the classification supervision was shown effective in generalizing the synthetic data-trained classifier to recognize measured data \cite{inkawhich2021bridging}. To address this, we investigate the relationships between generalization and transferability in the synthetic-to-measured recognition challenge \cite{lewis2019sar} and transfer attack.

	Augmenting training with Gaussian noise and changing model construction with dropout layers have proven quite effective in generalizing classifiers to process measured data after being trained with only synthetic data \cite{inkawhich2021bridging}. We compared these methods with our TEA, and a comparison of the average ASR and accuracy (ACC) results are listed in Table \ref{generalization}. With data augmentation and the addition of dropout layers, there were positive effects on generalization but negative effects on transferability. The Gaussian noise augmentation provided the best synthetic-to-measured recognition ACC but gave the worst average ASR. Our FT, which aligns the gradient w.r.t. the Gaussian noise augmented data, exhibited both positive results to generalization and transferability, but the key outcome is shown for AS enhancement, where the best transferability and worst generalization occurred simultaneously. This result shows that good generalization does not ensure strong transferability, and vice versa. The transferability may not be easily achieved by pursuing better generalization. Instead, there must be a balance between the two, and our TEA serves as one feasible solution. We can assume that aligning both the gradient and classification supervision may result in better generalization ability, but we leave that investigation for future studies.
	
	\begin{table}[tbp]
		\centering
		\caption{Average ASR and ACC  (\%) of RN18 with different methods. All methods were implemented with our FT process, and results are averaged over 5 trials. The $\sigma$ for Gaussian noise and drop rate $\tau$ for dropout layers were selected based on performance according to results reported in Fig. \ref{guasaug_dropout}.}
		\label{generalization}
		\begin{tabular}{lccccc}
			\toprule
			\multicolumn{1}{c}{} & \multicolumn{5}{c}{Model}  \\ \cmidrule{2-6}
			&  RN18  & +Gaus.  & +Dropout  & \cellcolor[rgb]{0.9,0.9,0.9}+FT  & \cellcolor[rgb]{0.9,0.9,0.9}+FT+AS \\ \midrule
			ACC & 66.47 & 90.91$\uparrow$ & 81.13$\uparrow$ & \cellcolor[rgb]{0.9,0.9,0.9}77.37$\uparrow$ & \cellcolor[rgb]{0.9,0.9,0.9}46.82$\downarrow$  \\
			ASR & 24.97 & 17.56$\downarrow$ & 20.86$\downarrow$ & \cellcolor[rgb]{0.9,0.9,0.9}36.24$\uparrow$ & \cellcolor[rgb]{0.9,0.9,0.9}45.22$\uparrow$  \\
			\bottomrule
		\end{tabular}
	\end{table}

	\subsection{Physical applicability of this study}\label{physicalapplicability}
\begin{figure}[tbp]
		\centering
		\includegraphics[width=1\linewidth]{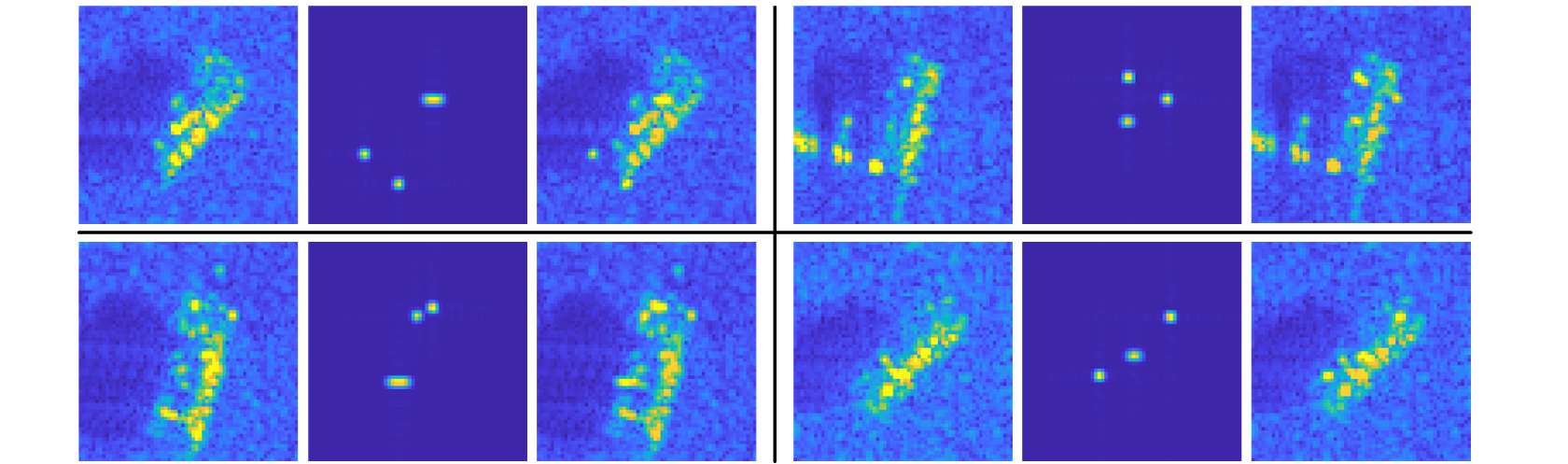}
		\caption{Adversarial examples generated by RN18+TEA and SMGAA with three adversarial scatterers. Each subplot group shows the original image, adversarial scatterers, and adversarial example from left to right.}
		\label{smgaaAE}
	\end{figure}

	\begin{table}[htb]
		\centering
		\caption{Results of SMGAA with the S2M attack setting \cite{peng2022scattering}. For 100 test images (10 for each class), we calculated 3 adversarial scatterers based on the surrogate model and synthetic data and transferred the resulting scatterers to the measured data against the target models. The best results are in \textbf{bold}.}
		\label{SMGAA}
		\begin{tabular}{l|cccc|c}
			\toprule
			\multicolumn{1}{c}{} & \multicolumn{4}{c}{Victim Model} & \\ \hline 
			\makecell[c]{Surr.} 	&  ACN  & SNV2  & RN18  & VGG16  & \textbf{Avg.} \\ \midrule 
			RN18 & 31 & 28 & 21 & 25 &  26.25 \\ 
			\cellcolor[rgb]{0.9,0.9,0.9}+TEA & \cellcolor[rgb]{0.9,0.9,0.9}\textbf{40} & \cellcolor[rgb]{0.9,0.9,0.9}\textbf{35}  & \cellcolor[rgb]{0.9,0.9,0.9}\textbf{28}  & \cellcolor[rgb]{0.9,0.9,0.9}\textbf{38}  & \cellcolor[rgb]{0.9,0.9,0.9}\textbf{35.25}  \\ 
			\bottomrule
		\end{tabular}
	\end{table}
	Although our main focus in this paper is on transferability in a more practical attack setting, this study naturally stays in line with the mainstream research in revealing the adversarial vulnerability by pursuing physical applicability. To illustrate, we show the compatibility of our method with the current physical-relevant studies. Current physical implementations of adversarial examples against SAR ATR can be divided into two categories: 1) implementing digital perturbations in the electromagnetic environment using jamming tools like phase-switched screen (PSS) \cite{xia2022sar} and 2) constraining the perturbations as parametric scattering centers \cite{qin2023scma,zhou2023attributed,peng2022scattering}. The digital attacks performed in Section \ref{experiment} could be implemented under the first category of physical attack.    Therefore, we experimented with the scattering model guided adversarial attack (SMGAA) \cite{peng2022scattering} to examine whether the TEA cooperates with scattering center-based methods. Table \ref{SMGAA} shows that with the TEA, the average ASR against three target models improved from 26.25\% to 35.25\% using the approach illustrated in Fig. \ref{smgaaAE}, where three extra adversarial scatterers were applied. Based on these results, the proposed TEA can cooperate with current physical-relevant research and help assess the adversarial risks in the practical S2M setting.

	\subsection{S2M variations}\label{s2mvariantions}
\begin{figure}[tbp]
		\centering
		\includegraphics[width=1\linewidth]{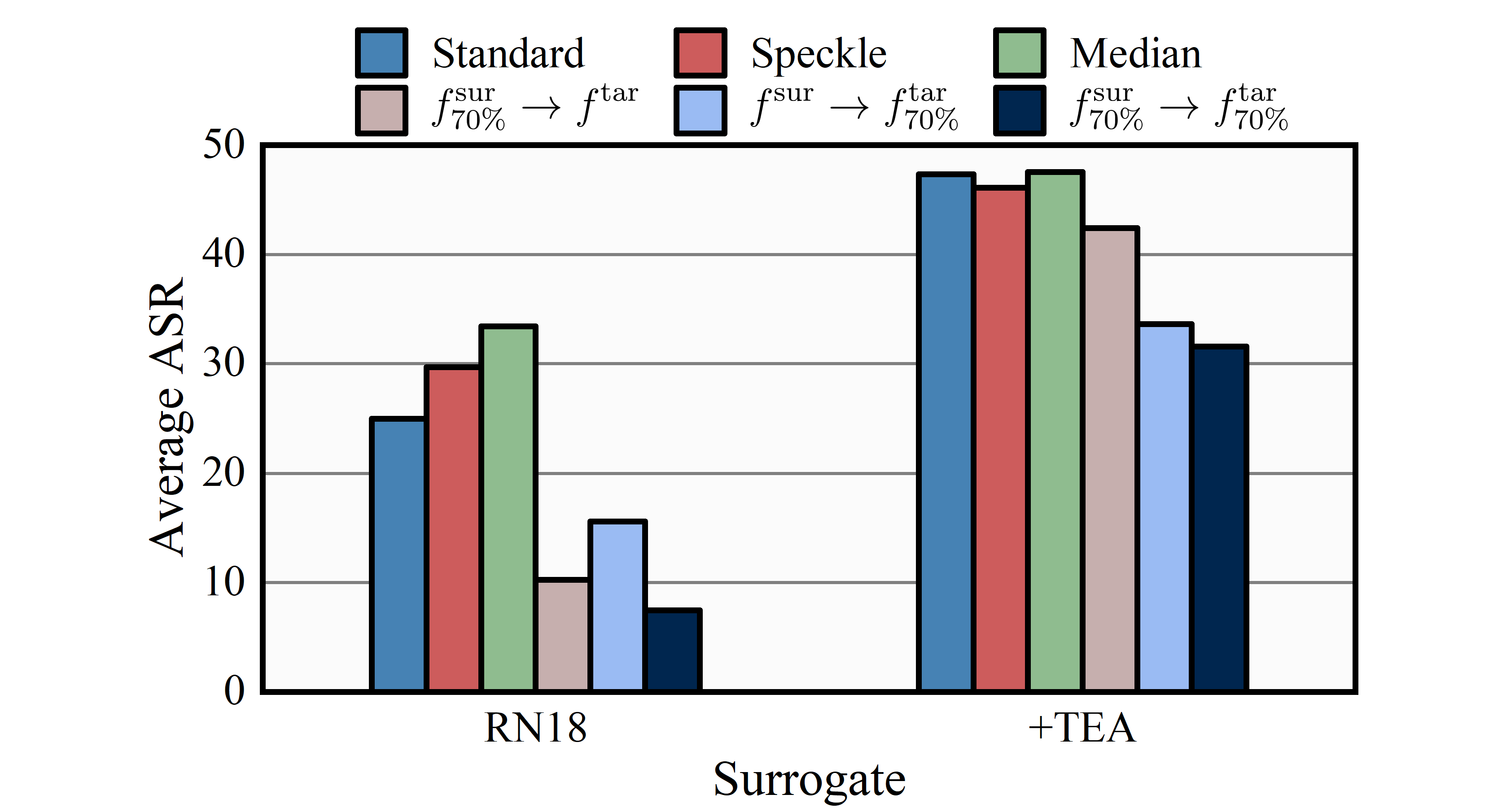}
		\caption{Average ASR (\%) against eleven target models. Speckle indicates that the surrogate models were trained over the synthetic data with multiplicative exponential distributed noise, and Median indicates that the training data was blurred by median filtering.}
		\label{s2mvariationsfig}
	\end{figure}

	In this section, we study the crucial factors that may affect the SAMPLE dataset-based S2M experiments, including the quality of synthetic data and the distribution mismatch between training data for surrogate and victim models. We considered two settings, speckle noise and median blur, as substitutes for degradation in data synthesis, and for training data mismatch, we further trained surrogate and victim models on random subsets that contained 70\% of the original synthetic and measured training data, respectively denoted as $f^{\text{sur}}_{70\%}$ and $f^{\text{tar}}_{70\%}$, and Fig. \ref{s2mvariationsfig} shows the average ASR results with the above settings. The results show that the two quality degradation cases studied had little effect on TEA with slight improvements over baseline instead of corruption. This may suggest that the SAMPLE synthetic data is not perfectly suited for S2M surrogate training, and the key quality factor affecting the performance is the electromagnetic structure of the target. Therefore, we may be able to further relax the restrictions of data synthesis. In contrast, the transferability suffered from the mismatch between training data distributions, while our TEA exhibited stable improvements in these settings. It is worth noting that the distribution mismatch challenge also exists in current MSTAR dataset-based M2M experiments, and the limited data capacity of SAR datasets could be a critical factor in this problem.
	
	\section{Conclusion and future work}\label{conclusion}
	Over the last few years, the adversarial vulnerability of DNN-based SAR ATR models has only been lightly explored, particularly in the setting where the victim's data is accessible. In this paper, we proposed a more practical S2M attack setting where attackers can only utilize synthetic data for designing adversarial perturbations, and we investigated potential threats under the S2M setting and proposed the TEA method. Without accessing the target data and model, the TEA can blindly enhance the S2M transferability of surrogate models and boost the aggressiveness of various attack algorithms, and our results indicate significant improvement in the gap between attacks with and without access to measured data. Overall, we shed new light on the adversarial vulnerability for SAR ATR, and our work highlights the urgent need to understand and secure ATR models in light of their vulnerability to adversarial attacks. 
	
	\color{black}{The next natural step to continue this work is to impose additional restrictions on the attacker, and these restrictions could include consideration of mismatches in the imaging algorithm, the imaging setting, observation geometries, and object categories between synthetic and measured data. Another potential research path is exploring transferability against advanced DNN inferences that incorporate scattering information. We also expect the proposed method to generalize to other ATR applications, such as high-resolution range profile \cite{du2022practical}, inverse SAR, and time-frequency features \cite{ma2023concealed}.
		
		\section{Declaration of Competing Interest}
		The authors declare that they have no known competing financial interests or personal relationships that could have appeared to influence the work reported in this paper.
		
		\section{Acknowledgments}
		This work was supported partially by the National Key Research and Development Program of China under Grant 2021YFB3100800, the National Natural Science Foundation of China under Grant 62376283, 61921001, 62022091, and 62201588, the Changsha Outstanding Innovative Youth Training Program under Grant kq2107002, and the Hunan Graduate Research Innovation Project under Grant CX20230044.
		
		\bibliographystyle{cas-model2-names}
		
		\bibliography{refediffasc}

	\end{document}